\documentclass{article}
\usepackage{amsmath}

\PassOptionsToPackage{numbers, compress}{natbib}

\usepackage[preprint]{neurips_2025}

\usepackage[utf8]{inputenc} %
\usepackage[T1]{fontenc}    %
\usepackage{hyperref}       %
\usepackage{url}            %
\usepackage{booktabs}       %
\usepackage{amsfonts}       %
\usepackage{nicefrac}       %
\usepackage{microtype}      %
\usepackage{xcolor}         %
\usepackage{graphicx}
\usepackage{subcaption}
\usepackage{booktabs}
\usepackage{tabularx}
\usepackage{epsfig}
\usepackage{graphicx}
\usepackage{natbib}
\usepackage{amsmath}
\usepackage{amssymb}
\usepackage{colortbl}
\usepackage{multirow}
\usepackage{xspace}
\usepackage{color}
\usepackage{setspace}
\usepackage[normalem]{ulem}
\usepackage{tikz}
\usepackage{comment}
\usepackage{algorithm}
\usepackage{algpseudocode}
\usepackage{float}
\usepackage{caption}
\usepackage{adjustbox}
\usepackage{pifont}%
\usepackage{wrapfig}
\usepackage{booktabs,tabularx,makecell}
\newcommand{\cmark}{\ding{51}}%
\newcommand{\xmark}{\ding{55}}%

\definecolor{top1}{RGB}{245,152,153}
\definecolor{top2}{RGB}{253,205,154}
\definecolor{top3}{RGB}{248,244,140}

\usepackage[capitalize]{cleveref}
\crefname{section}{Sec.}{Secs.}
\Crefname{section}{Section}{Sections}
\Crefname{table}{Table}{Tables}
\crefname{table}{Tab.}{Tabs.}

\usepackage{enumitem}

\def\ourmodel{\texttt{FreSca}\xspace} 

\title{FreSca: Scaling in Frequency Space Enhances Diffusion Models}

\author{Chao Huang\textsuperscript{1}, Susan Liang\textsuperscript{1},  Yunlong Tang\textsuperscript{1}, Jing Bi\textsuperscript{1}, Li Ma\textsuperscript{2}, Yapeng Tian\textsuperscript{3}, Chenliang Xu\textsuperscript{1} \\
\textsuperscript{1}University of Rochester,  \textsuperscript{2}HKUST, \textsuperscript{3}The University of Texas at Dallas\\
}

\begin{document}

\maketitle

\begin{center}
\includegraphics[width=0.99\linewidth]{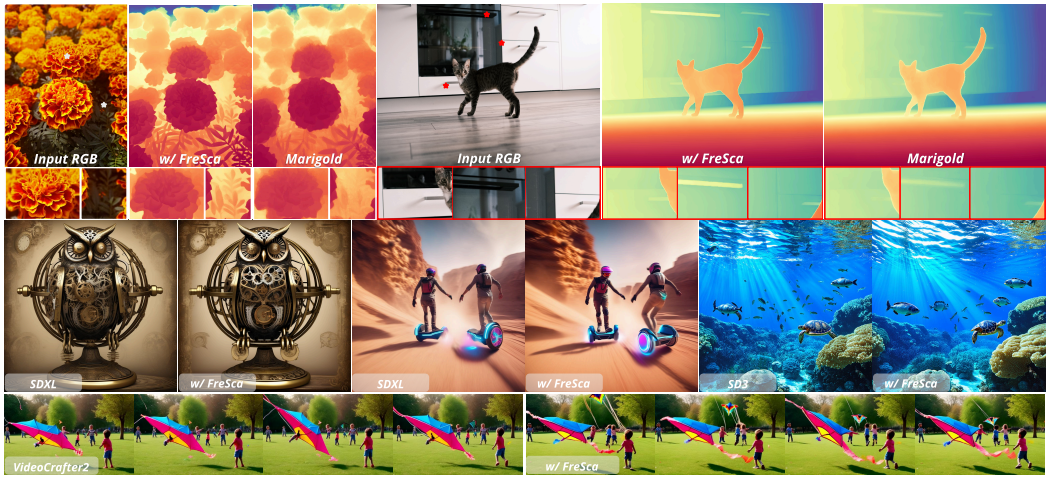}
\captionof{figure}{\textbf{\ourmodel: A plug-and-play enhancement for diffusion models.} Without retraining, \ourmodel refines Marigold~\cite{ke2023repurposing} depth predictions to recover fine details (top); enables precise, prompt-aligned generation over SD3~\cite{sd3} (middle) ; and boosts motion, detail, and temporal consistency in VideoCrafter2~\cite{chen2024videocrafter2} video generation (bottom) .}
\label{fig:teaser}
\end{center}
  
\begin{abstract}

Latent diffusion models (LDMs) have achieved remarkable success in a variety of image tasks,  yet achieving fine-grained, disentangled control over global structures versus fine details remains challenging. This paper explores frequency-based control within latent diffusion models. We first systematically analyze frequency characteristics across pixel space, VAE latent space, and internal LDM representations. This reveals that the ``noise difference'' term, $\Delta\epsilon_t$, derived from classifier-free guidance at each step $t$, is a uniquely effective and semantically rich target for manipulation.  Building on this insight, we introduce {\ourmodel}, a novel and plug-and-play framework that decomposes $\Delta\epsilon_t$ into low- and high-frequency components and applies independent scaling factors to them via spatial or energy‐based cutoffs. Essentially, {\ourmodel} operates without any model retraining or architectural change, offering model- and task-agnostic control. We demonstrate its versatility and effectiveness in improving generation quality and structural emphasis on multiple architectures (e.g., SD3, SDXL) and across applications including image generation, editing, depth estimation, and video synthesis, thereby unlocking a new dimension of expressive control within LDMs.

\end{abstract}

\section{Introduction}

Latent diffusion models (LDMs)~\citep{rombach2022high} have emerged as a dominant force in generative modeling, capable of producing images of unprecedented quality and diversity from textual prompts~\cite{sdxl,sd3,dalle2,imagen} or other conditioning signals~\cite{controlnet}. Despite their power, achieving nuanced control beyond the initial conditioning remains an active area of research. Users often desire to modulate specific image characteristics, such as the prominence of fine textures versus coarse shapes, or to impart particular artistic styles, in a more direct and disentangled manner. Existing control mechanisms might involve complex model modifications, additional training, or offer only coarse‐grained adjustments.

The frequency domain offers a natural and powerful paradigm for image manipulation~\cite{adelson1984pyramid}, where low frequencies typically represent global structures and smooth variations, while high frequencies encode fine details such as edges and textures. This fundamental separation has been exploited in classical image processing for tasks like sharpening~\cite{gastal2011domain}, denoising~\cite{ergen2012signal}, and style transfer~\cite{deng2019wavelet}. We hypothesize that by extending frequency‐domain manipulations to the internal workings of LDMs, we can unlock more intuitive and fine‐grained control over the synthesis process. However, the iterative nature of diffusion and its operation within a learned noisy latent space raise critical questions: How do frequency characteristics translate from pixel space to the VAE latent space? And, more importantly, which specific component or stage within the diffusion model’s denoising trajectory is most amenable and effective for frequency‐based interventions?

\begin{figure}[t]
    \centering
    \includegraphics[width=0.99\linewidth]{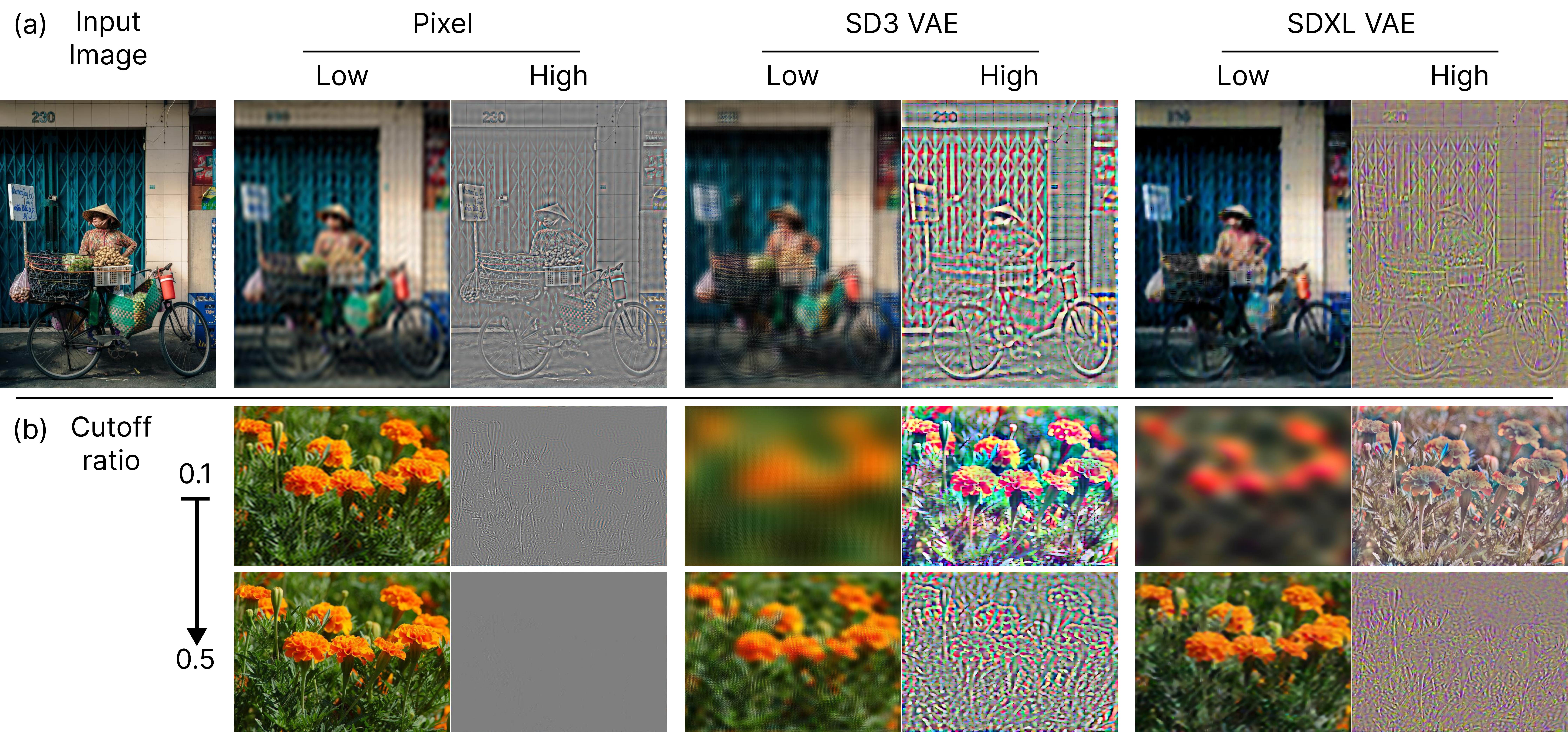}
    \caption{(a) Frequency decomposition of an RGB image $(I_l,I_h)$ and its SD3~\cite{sd3}/SDXL~\cite{sdxl} VAE encodings $(x_l,x_h)$ with $r_0=0.05$ (pixel) and $r_0=0.5$ (latent). (b) Cutoff‐radius sensitivity in pixel vs.\ latent space.}
    \label{fig:vae_pixel}
\end{figure}

In this paper, we systematically investigate these questions. We begin by comparing frequency decompositions in pixel space versus the VAE latent space (as shown in \cref{fig:vae_pixel}), highlighting differences in semantic content and sensitivity. Grounded by the observations on the VAE latent space, we then explore various internal representations within the diffusion model, including the noisy latents $\mathbf{x}_t$, the noise prediction $\epsilon_t$, and the crucial ``noise difference'' term $\Delta\epsilon_t$ arising from classifier‐free guidance (CFG)~\citep{ho2022classifier}. Interestingly, our analysis reveals that $\Delta\epsilon_t$ is particularly rich in semantic information among others, and  thereby can serve as an ideal target for frequency manipulation.

Based on these insights, we propose \texttt{\ourmodel}, a versatile framework that operates by decomposing the noise prediction into its low‐ and high‐frequency components at each step of the denoising process. \texttt{\ourmodel} then applies distinct scaling factors to these components, allowing for independent amplification or suppression of global structures and fine details. To further enhance adaptability, \texttt{\ourmodel} supports both spatial- and energy-based frequency cutoffs for band separation. As \texttt{\ourmodel} operates directly in the common noise space used by nearly all diffusion models, it is inherently model- and task-agnostic, avoiding the architectural constraints of prior frequency-aware methods~\citep{si2023freeu,bu2024broadway}. We validate this versatility across a variety of models (e.g., SDXL~\cite{sdxl}, SD3~\cite{sd3}) and tasks such as diffusion-based depth estimation~\cite{ke2023repurposing,shao2024learning}, image generation~\cite{sd3,sdxl}, image editing~\cite{brack2024ledits++,huberman2024edit}, and video synthesis~\cite{chen2024videocrafter2}.

In summary, our contributions to the community are
\begin{enumerate}[leftmargin=1.5em]
  \item A comparative analysis of frequency representations in pixel space, VAE latent space, and key internal states of latent diffusion models.
  \item The identification of the CFG‐derived noise‐difference term, $\Delta\epsilon_{t}$, as a highly effective and semantically meaningful target for frequency‐based manipulation in LDMs.
  \item The \texttt{\ourmodel} framework, a plug-and-play method providing disentangled control over low- and high-frequency image characteristics without model retraining or architectural changes. We demonstrate its efficacy through qualitative and quantitative experiments on diverse tasks and models, highlighting its ability to produce varied stylistic effects and modulate detail levels.
\end{enumerate}

\section{Related Works}

\noindent\textbf{Controls in Diffusion Models.}
The quest for greater control over diffusion model outputs has spurred various approaches. Prompt engineering~\cite{witteveen2022investigating} is the most direct method but often lacks fine-grained control over specific visual attributes. Classifier-Free Guidance~\citep{ho2022classifier} significantly improved sample quality and adherence to prompts by amplifying the guidance signal.
Beyond prompt-based control, structural guidance methods like ControlNet~\citep{controlnet} and T2I-Adapters~\citep{mou2024t2i} enable conditioning on spatial inputs like edge maps or pose, typically by introducing trainable modules or fine-tuning parts of the U-Net. Other approaches focus on adapting pre-trained models using lightweight finetuning techniques such as LoRA~\citep{lora} for domain-specific generation.
While powerful, many of these methods may require auxiliary networks~\cite{controlnet,mou2024t2i}, per-instance optimization~\cite{null-text,dreambooth,text-inversion}, or are not primarily focused on disentangled frequency control. In contrast, \texttt{\ourmodel} differs by offering a zero-shot, plug-and-play mechanism that directly targets frequency bands during the denoising process of diffusion models.

\noindent\textbf{Frequency and Spectral Methods in Diffusion Models.}
Frequency and spectral analyses have long illuminated deep models’ behavior, from CNNs’ spectral bias~\citep{rahaman2019spectral} to distribution discrepancies in GANs~\citep{durall2020watch}. Yet, despite these insights and analogous explorations in neural networks, explicit frequency-domain control within diffusion processes remains nascent. A handful of recent works have sought to manipulate spectral components, e.g., tuning the frequency behavior of U-Net skip connections and backbone features~\citep{si2023freeu}, applying filters to noisy latents for artistic effects~\citep{geng2024visual,geng2024factorized}, and modulating frequency content in temporal attention maps~\citep{bu2024broadway}. However, these approaches tend to be model- or task-specific and do not generalize across diffusion variants. In contrast, \texttt{\ourmodel} offers a unified, model- and task-agnostic framework to decompose and dynamically scale the classifier-free guidance noise difference by frequency, providing direct, interpretable control over both global structure and fine detail.

\section{Method}

In this section, we begin by analyzing the differences between frequency decomposition in pixel space versus the latent space of Variational Autoencoders (VAEs). Subsequently, we investigate frequency decomposition applied to various intermediate representations within diffusion models to pinpoint an effective basis for frequency manipulation. We then examine the denoising trajectory, observing the step-wise dynamics of different frequency bands. Building on these insights, we introduce \texttt{\ourmodel}, a novel framework for unified frequency scaling in latent diffusion models.

\noindent\textbf{Preliminaries.}
Latent diffusion models (LDMs) operate by first encoding images into a latent space using a VAE, and then performing the diffusion process within this space. An LDM typically consists of: (i) a VAE with an encoder $\mathcal{E}$ and a decoder $\mathcal{D}$. Given an RGB image $I$,  the encoder maps it to an initial latent representation $\mathbf{x} = \mathcal{E}(I)$. The decoder reconstructs the image from a latent code as $\hat{I} = \mathcal{D}(\mathbf{x})$. (ii) A time-conditional denoising network $\epsilon_{\theta}$ that operates in the latent space. 
The diffusion model involves a forward noising process and a reverse denoising process over $T$ timesteps. 
Starting with an initial latent $\mathbf{x}_0$, the forward process corrupts it into a squence of noisy latent $\{\mathbf{x}\}_{t=1}^T$ by gradually adding Gaussian noise according to predefined schedule (see, e.g.,~\citep{ho2020denoising}). At each time step $t$, the denoising network $\epsilon_{\theta}(\mathbf{x}_t, t)$ is trained to predict the added noise $\epsilon_t$, enabling a reverse denoising process that recovers $\mathbf{x}_0$ from pure noise. In what follows, $\mathbf{x}_t$ denotes the latent at timestep $t$, and $\epsilon_{\theta}$ the noise predictor--our primary handle for frequency-based control.

\subsection{Frequency Decomposition in Pixel vs.\ Latent Space}
Frequency decomposition is a cornerstone of image processing, enabling insights into both classical algorithms and modern neural networks. Typically, an image can be separated into low-frequency components, capturing global structures and smooth variations, and high-frequency components, encoding fine details like edges and textures. 

While this concept is well-established in pixel space, its extension to the latent representations learned by VAEs (and subsequently used by LDMs) requires investigation. To this end, we define a unified frequency decomposition operator. Given an input signal 
$
u \;\in\;\{\,I\;\text{(RGB image)},\;\mathbf{x}\;\text{(VAE latent)}\},
$
we compute its channel-wise 2D Fourier transform:
\begin{equation}    
U \;=\; \mathcal{F}(u),
\quad
u \;=\; \mathcal{F}^{-1}(U).
\label{eq:transform}
\end{equation}
Let the spatial dimensions of $u$ be $H\times W$. 
We define a cutoff ratio $r_0\in[0,1]$, which the actual cutoff radius $R_c$ in the frequency domain is then $R_c = r_0 \cdot \min(H/2, W/2)$. This ensures the ratio $r_0$ has comparable effect across different spatial resolutions. We then define binary low-pass $M_l$ and high-pass $M_h$ masks over frequency coordinates $(k_x,k_y)$:
\begin{equation}
M_l(k_x, k_y)  \;=\;
\begin{cases}
1, & \text{if } \sqrt{k_x^2 + k_y^2} \le R_c,\\
0, & \text{otherwise} ,
\end{cases}
\quad
M_h(k_x, k_y) = 1 - M_l(k_x, k_y).
\label{eq:thresholding}
\end{equation}

The low- and high-frequency components of \(u\) are then obtained by applying these masks in the Fourier domain:
\begin{equation}
  u_l \;=\; \mathcal{F}^{-1}\bigl(M_l \,\odot\, U\bigr),
  \quad
  u_h \;=\; \mathcal{F}^{-1}\bigl(M_h \,\odot\, U\bigr),
\label{eq:masking}
\end{equation}
with \(\odot\) denoting element-wise multiplication.

By applying this decomposition to both the pixel image $I$ and its VAE encoding $\mathbf{x}$, we obtain pairs $(I_l,I_h)$ and $(x_l,x_h)$. 
Visual comparisons (see \cref{fig:vae_pixel}(a)) indicate that in both domains, low frequencies correspond to coarse structures and high frequencies to details. However, we identify two key distinctions: (i) \underline{\textit{Semantic richness in latent high frequencies.}}  The high-frequency components of $\mathbf{x}$ tend to preserve more abstract semantic patterns, such as object contours and characteristic textures. This reflects the VAE's ability to learn meaningful representations. (ii) \underline{\textit{Threshold sensitivity.}} Pixel-space details (edges, textures) diminish rapidly as $r_0$ increases (e.g., beyond 0.1). In contrast, VAE latent features often reveal significant structural and textural information even at higher $r_0$ value (see \cref{fig:vae_pixel}(b)).
These observations highlight both the conceptual alignment and the practical differences in frequency content between pixel and VAE latent spaces.

\begin{table}[!t]
  \centering
  \caption{Experiment configuration: Frequency operations (\cref{eq:transform,eq:thresholding,eq:masking} ) applied across different feature spaces.}
  \label{tab:frequency_ops}
  \begin{tabularx}{\textwidth}{@{}l c c *{3}{>{\centering\arraybackslash}X}@{}}
    \toprule
    \textbf{Operation} 
      & \textbf{Pixel} 
      & \textbf{VAE} 
      & \multicolumn{3}{c}{\textbf{Diffusion Model Space}} \\
    \cmidrule(lr){4-6}
      & 
      &  
      & Noisy Latents  
      & Combined Noise  
      & Noise Difference \\
    \midrule
    \cref{eq:transform,eq:thresholding,eq:masking} 
      & $I$ 
      & $\mathbf{x}$ 
      & $\mathbf{x}_{1:T}= \{\mathbf{x}_t\}_{t=1}^T$ 
      & $\epsilon_{1:T}=\{\epsilon_t\}_{t=1}^T$ 
      & $\Delta\epsilon_{1:T} = \{\Delta\epsilon_t\}_{t=1}^T$ \\
    \bottomrule
  \end{tabularx}
\end{table}

\subsection{Frequency Decomposition for Diffusion Models}
Having analyzed frequency characteristics in the VAE latent space, we now investigate where frequency-specific manipulations can be most effectively applied during the iterative denoising process of LDMs. For conditional generation (e.g., text-to-image), LDMs typically employ Classifier-Free Guidance~\citep{ho2022classifier}. The effective noise prediction $\epsilon_{t}$ at timestep $t$ is:
\begin{equation}
\label{eq:cfg}
    {\epsilon}_t = \epsilon_\theta(\mathbf{x}_t, t) + \omega \cdot \Delta\epsilon_t, \quad \Delta\epsilon_t = \epsilon_\theta(\mathbf{x}_t, {\boldsymbol{c}}, t) - \epsilon_\theta(\mathbf{x}_t, t).
\end{equation}

Here $\epsilon_\theta(\mathbf{x}_t, {\boldsymbol{c}}, t)$ and $\epsilon_\theta(\mathbf{x}_t, t)$ denote the conditional and unconditional noise estimates, and $\omega$ is the classifier-free guidance scale.

\begin{figure}[!t]
    \centering
    \includegraphics[width=0.99\linewidth]{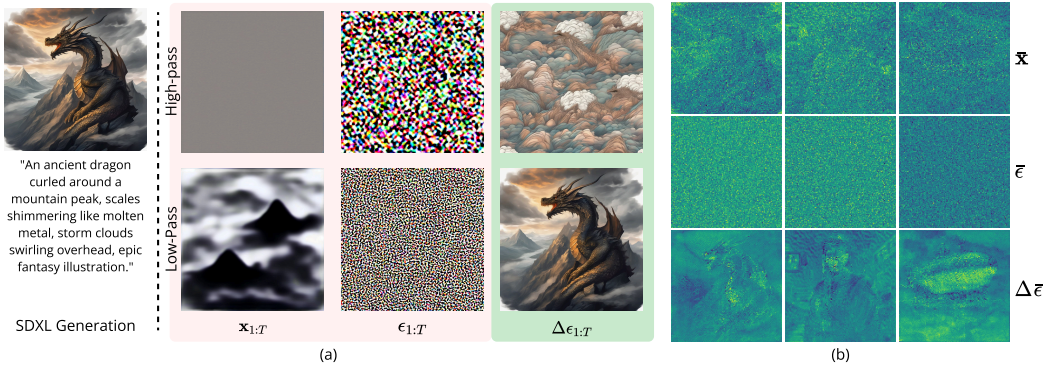}
    \caption{(a) SDXL outputs (left) and results of frequency decomposition on various diffusion representations (right); top: high‐frequency components, bottom: low‐frequency components; cutoff $r_0=0.5$. (b) Temporal average over $T$ steps for each representation, highlighting the semantic richness of the noise‐difference term.}
    \label{fig:representation}
\end{figure}
\begin{figure}[!t]
    \centering
    \includegraphics[width=0.99\linewidth]{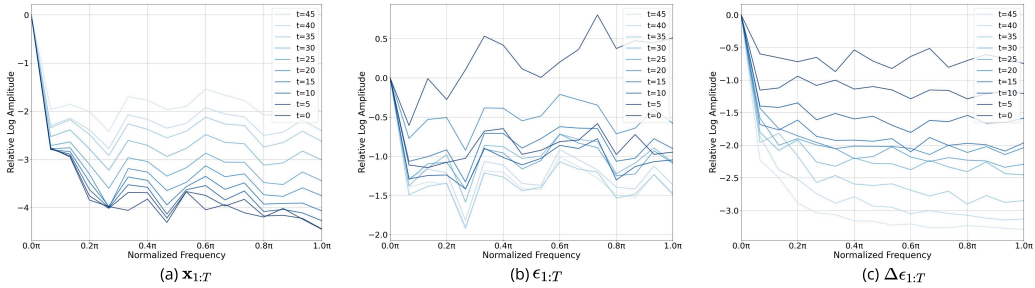}
    \caption{Relative log amplitudes of Fourier over all $T$ denoising steps for (a) the latent variables $\mathbf{x}_t$, (b) the noise prediction $\epsilon_t$, and (c) the noise‐difference term $\Delta\epsilon_t$. Each curve corresponds to a timestep, illustrating how low and high frequencies changes in each representation.}
    \label{fig:step}
\end{figure}

We consider three primary candidate representations within the diffusion process for applying frequency decomposition (using \cref{eq:transform,eq:thresholding,eq:masking}), as outlined in \cref{tab:frequency_ops}. To determine the most suitable candidate, we apply either a low-pass or a high-pass filter (using a fixed $r_0=0.5$) to the chosen representation at each denoising step $t$.
The final generated image $\mathcal{D}(\mathbf{x}_0)$ allows us to assess the impact. 
Our experiments (visualized in \cref{fig:representation}(a)) reveal that manipulating the frequency components of the noise difference term $\Delta\epsilon_{1:T}$ yields the most semantically meaningful and controllable results. For instance, removing high-frequency components from $\Delta\epsilon_{1:T}$ results in minimal degradation to the overall image structure, while selectively preserving only its high-frequency components can produce interesting stylization effects, capturing low-level textures of patterns like ``dragon,'' ``cloud,'' and ``mountain.'' 

\begin{figure}[t]
  \centering
  \begin{minipage}{0.47\textwidth}
    \centering
    \includegraphics[width=\textwidth]{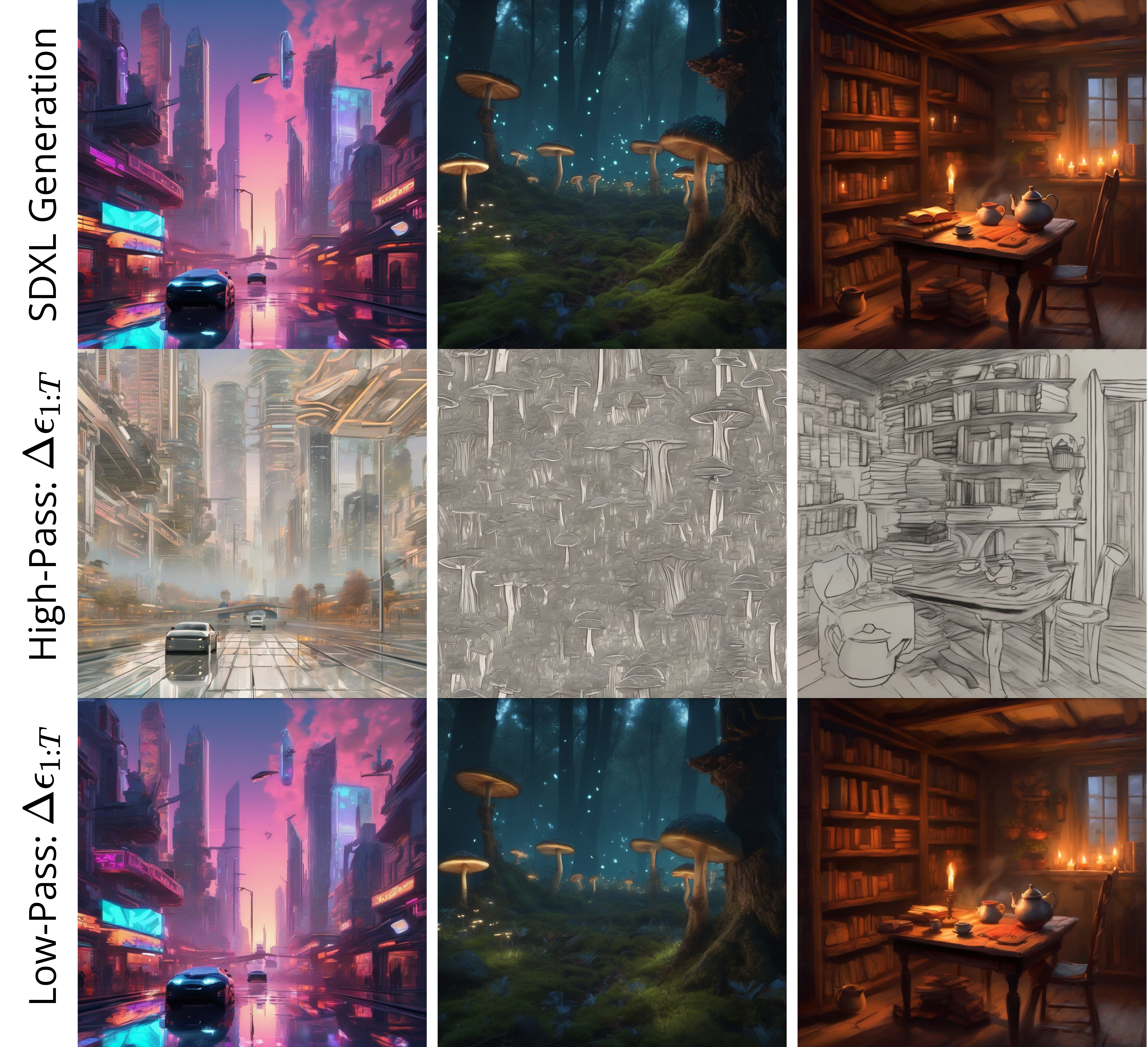}
    \captionof{figure}{Examples of original SDXL generations (top) and the generation results by applying high‐pass (middle) and low-pass filters (bottom) on $\Delta\epsilon_{1:T}$.}
    \label{fig:high_low}
  \end{minipage}\hfill
  \begin{minipage}{0.43\textwidth}
    \centering
    \includegraphics[width=\textwidth]{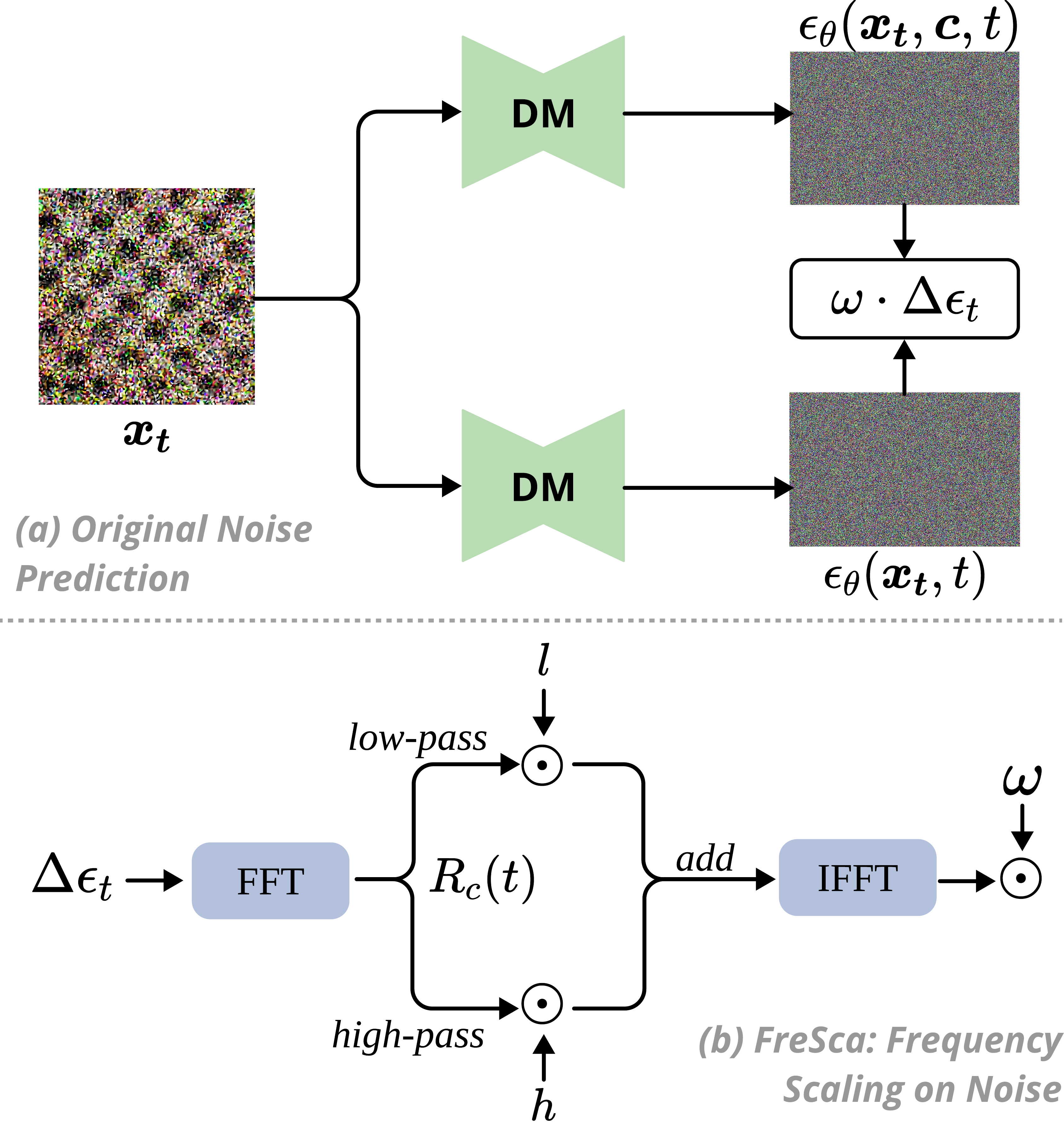}
    \captionof{figure}{Overview of \ourmodel. We introduce scaling factors $l$ and $h$ to decompose the control mechanisms in the Fourier domain.}
    \label{fig:method}
  \end{minipage}
\end{figure}

We hypothesize that $\Delta\epsilon_{1:T}$ inherently encodes crucial semantic structures. To support this, we normalize each of the three candidate sequences (per-channel min-max normalization at each step t) and then time-average them, yielding $\bar{\mathbf{x}}$, $\bar{\epsilon}$, and $\Delta\bar{\epsilon}$. As shown in \cref{fig:representation}(b), $\bar{\Delta\epsilon}$ exhibits clearer semantic structures compared to the others, suggesting it is a more potent target for frequency-based operations. Further examples in 
\cref{fig:high_low} corroborate the significant role of frequency components within $\Delta\epsilon_t$.

\noindent\textbf{Step-wise Frequency Dynamics.} Based on our analysis of the three diffusion representations, we further examine their evolution of spectral profiles throughout the denoising trajectory (see \cref{fig:step}). Our key observations are:
\begin{enumerate}[leftmargin=1.5em, itemsep=0.4em]
    \item The spectrum of the noisy latent $\mathbf{x}_t$ shows that low-frequency structures quickly converge in early step, and emerge more clear in later steps as the high-frequency noise is attenuated.
    \item The specturm of $\epsilon_{1:T}$ shows more flutations, and no consistent trend is found across different $t$.
    \item $\Delta\epsilon_{1:T}$ evolves from a more low-pass characteristic at early, high-noise stages towards a broader, flatter spectrum at later stages. Furthermore, as $t$ decreases, its magnitude generally increases, signifying that the guidance becomes more influential in refining details during later steps.
\end{enumerate}

\subsection{\texttt{\ourmodel}: Versatile Frequency Scaling in Diffusion Models}
Building on the finding that the noise difference term $\Delta\epsilon_{t}$ is a semantically rich and suitable candidate for frequency manipulation, we introduce \texttt{\ourmodel},  a framework for versatile frequency scaling within LDMs. \texttt{\ourmodel} operates by decomposing $\Delta\epsilon_{t}$ into its low- and high-frequency components and then applying independent scaling factors to each.

Let $U_t$ be the Fourier transform of the noise difference term at timestep $t$. Using the low-pass ($M_l$) and high-pass ($M_h$) masks defined in \cref{eq:thresholding} (which depend on a cutoff choice, see below), we define the modified noise difference term $\hat{\Delta\epsilon_t}$ as:
\begin{equation}
\hat{\Delta\epsilon_t} = \mathcal{F}^{-1}\bigl( l \cdot M_l \odot U_t + h \cdot M_h \odot U_t \bigr),
\label{eq:ourmodel_op}
\end{equation}
where we introduce two scaling factors $l$ and $h$ that allow for independent amplification or suppression of different frequency bands. This modified $\hat{\Delta\epsilon_t}$ then replaces ${\Delta\epsilon_t} $ in \cref{eq:cfg}. 
Generally, \texttt{\ourmodel} offers several advantages: 
\begin{itemize}[leftmargin=1.5em]
    \item \textbf{Flexibility:} Independent scaling of low and high frequencies enables effects from fine‐detail enhancement ($h>1,l=1$) to smoothing ($l>1,h<1$) or targeted stylization of specific bands.
    \item \textbf{Faithfulness:} When $l=h=1$, \texttt{\ourmodel} losslessly reduces to the original CFG mechanism.
    \item \textbf{Generality:} As it operates on the noise difference term, a ubiquitous component of conditional diffusion, {\ourmodel} applies seamlessly across architectures (e.g., SDXL, SD3) and tasks.
\end{itemize}

\noindent\textbf{Dynamic Cutoff Determination.} The effectiveness of {\ourmodel} can be further enhanced by dynamically adjusting the frequency separation (i.e., the cutoff radius $R_c$ used in $M_l,M_h$) at each timestep $t$. We propose two strategies for determining $R_c(t)$:
\begin{enumerate}[leftmargin=1.5em]
    \item \textbf{Spatial-Ratio Cutoff:} The cutoff radius $R_c(t)$ is determined based on a predefined ratio $r_0$:
    \begin{equation}
        R_c(t) = r_0 \cdot \min(H_t/2, W_t/2),
    \label{eq:spatial_ratio_cutoff}
    \end{equation}
    where $H_t$ and $W_t$ are the spatial dimension of $U_t$.
    
    \item \textbf{{Energy-Based Cutoff:} } 
    Let $E_{\rm tot}(t) \;=\;\sum_{k_x,k_y}\bigl|U_t(k_x,k_y)\bigr|$. We choose the smallest integer $R$ such that the cumulative magnitude within radius $R$ reaches a fraction $r_0$ of $E_{\rm tot}(t)$:
    \begin{equation}
    R_c(t)
    = \min\Bigl\{\,R\in\mathbb{N}_0 \;\mid\;
    \sum_{\sqrt{k_x^2+k_y^2}\,\le\,R}
    \bigl|U_t(k_x,k_y)\bigr|
    \;\ge\;r_0\,E_{\mathrm{tot}}(t)\Bigr\}\,.
    \label{eq:energy_cutoff}
    \end{equation}
    This tailors ``low'' versus ``high'' frequencies to the spectral energy distribution at each step.
\end{enumerate}

\begin{figure}[!t]
    \centering
    \includegraphics[width=1\linewidth]{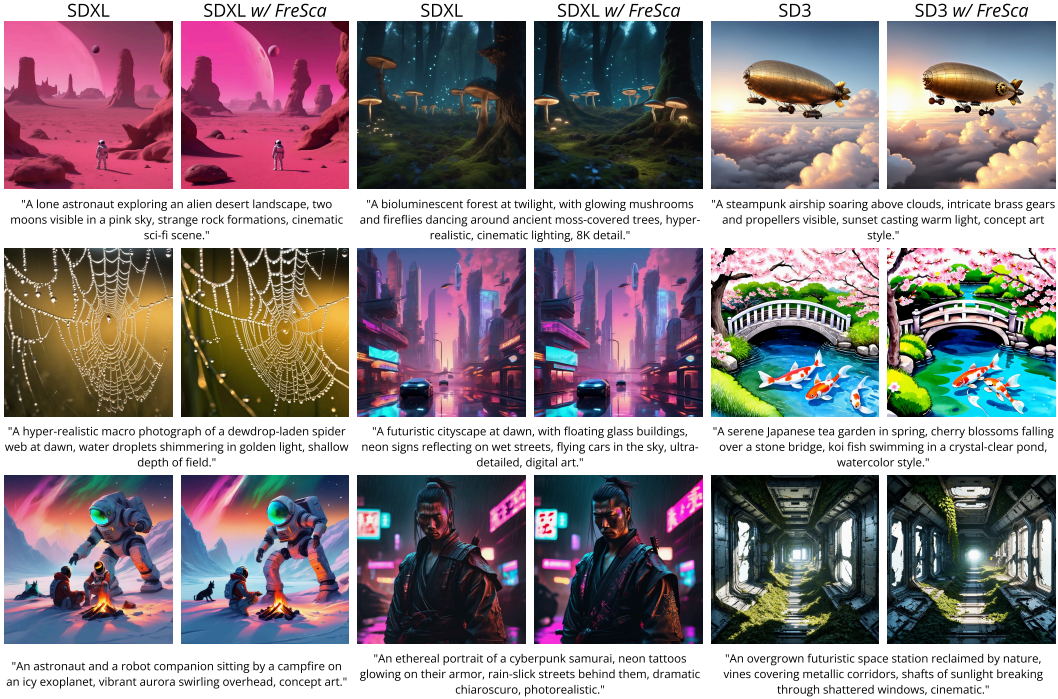}
    \caption{Samples generated by SDXL~\cite{sdxl} and SD3~\cite{sd3} with or without \ourmodel.}
    \label{fig:sd3_sdxl}
\end{figure}

\begin{figure}[t]
  \centering
  \begin{minipage}{0.7\textwidth}
    \centering
    \includegraphics[width=\textwidth]{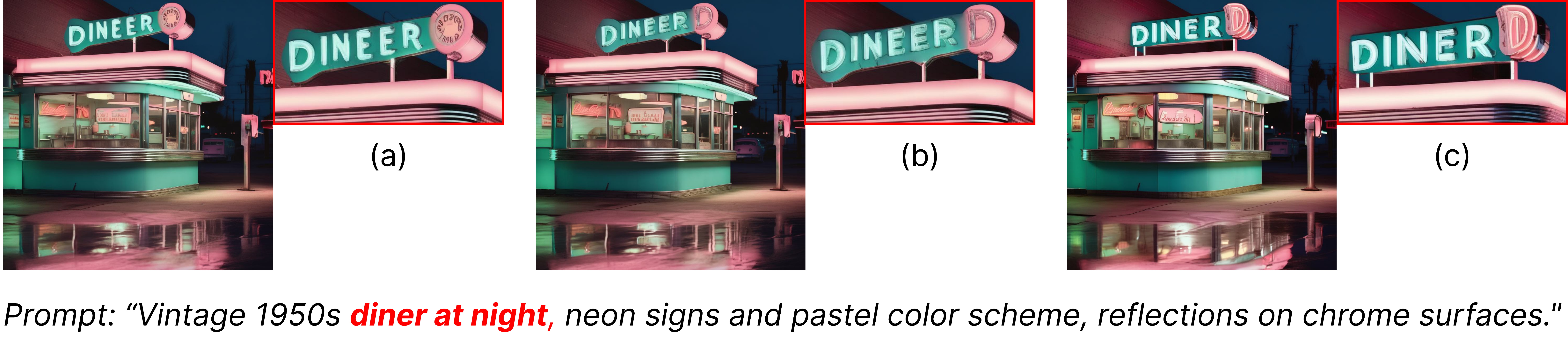}
    \captionof{figure}{Ablation of cutoff strategies: (a) original SDXL output; \ourmodel applied with (b) spatial‐ratio cutoff and (c) energy‐based cutoff (both $h=1.5$). The adaptive energy‐based cutoff yields the closest alignment to the prompt. }
    \label{fig:energy_versus_spatial}
  \end{minipage}\hfill
  \begin{minipage}{0.28\textwidth}
    \centering
    \includegraphics[width=\textwidth]{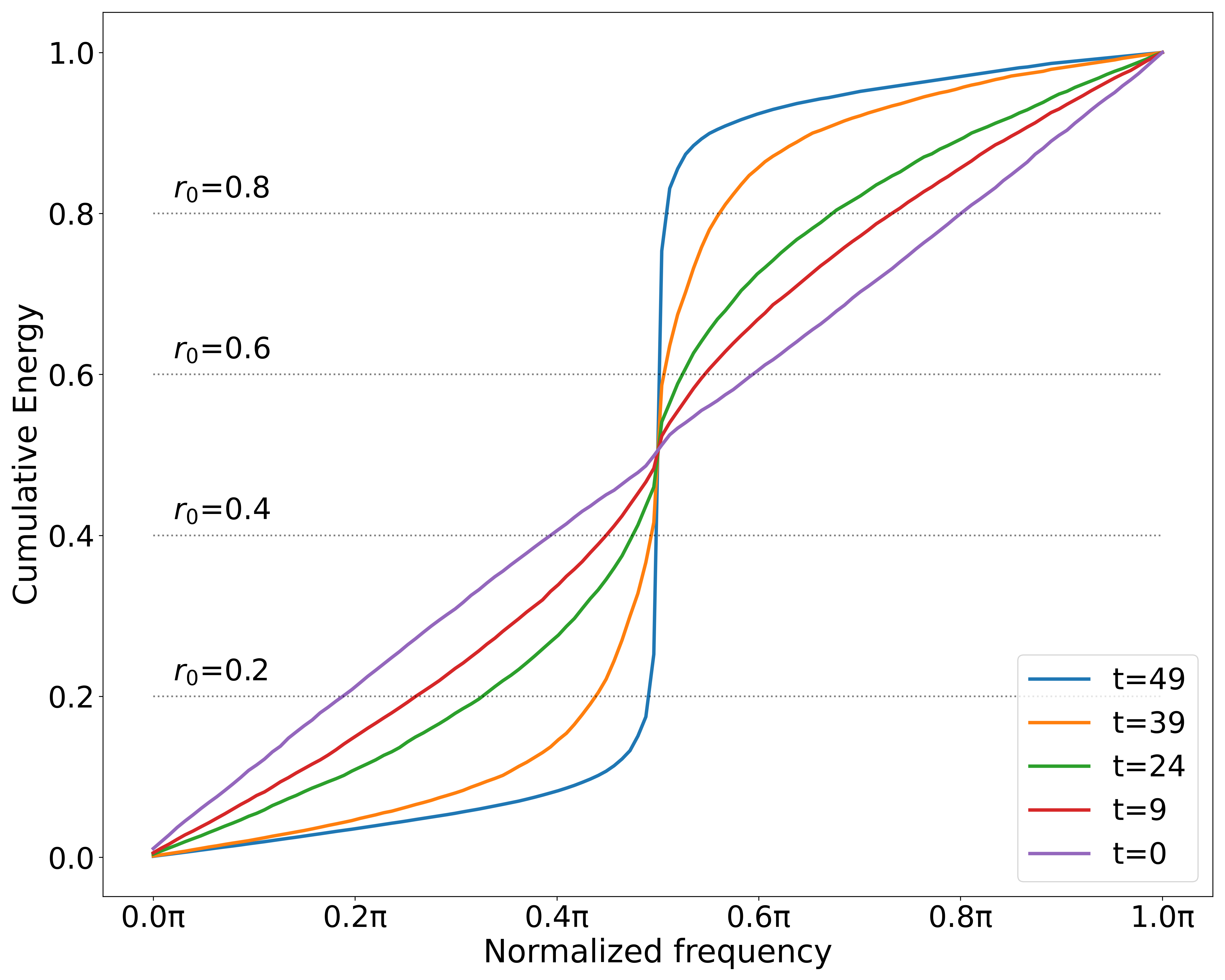}
    \captionof{figure}{Cumulative‐energy curve that tells how $r_0$ affects cutoff radius at timestep $t$.}
    \label{fig:energy_schedule}
  \end{minipage}
\end{figure}

\section{Experiments: \texttt{\ourmodel} is Vesatile, Model-Agnostic, and Task-Agnostic}

\subsection{Task: Text to Image Generation}
\noindent\textbf{Generalization Across Models.} To demonstrate \ourmodel’s model‐agnostic versatility, we incorporate it into two distinct image generation methods: SDXL~\cite{sdxl}, which uses a U‐Net backbone, and SD3~\cite{sd3}, a multimodal diffusion transformer. In \cref{fig:sd3_sdxl}, both setups employ a high‐frequency scaling factor $h=1.5$ with an energy‐based cutoff $r_0=0.9$. In each case, \ourmodel enhances prompt fidelity and overall image quality, producing outputs that better match the text description while exhibiting fewer distortions.

\noindent\textbf{Ablation on Cutoff Strategy.} In \cref{fig:energy_versus_spatial}, we compare the baseline SDXL output against \ourmodel using spatial‐ratio and energy‐based cutoffs. The energy‐based variant, with its adaptive radius schedule shown in \cref{fig:energy_schedule}, produces generations that more closely match the prompt.

More image generation results and ablations on the effect of scaling factors $h$, $l$, and different cutoff ratio can be found in the supplementary materials.

\begin{table}[!t]
    \centering
    \caption{\textbf{Zero-shot depth estimation on DIODE, KITTI, and ETH3D.} We compare Marigold and Marigold + \ourmodel using AbsRel (lower better) and $\delta_1$ (higher better); \textbf{bold} denotes best, \underline{underline} represents second best. Our method consistently improves both indoor and outdoor results. $^\dagger$Official Marigold implementation.}

    \label{table:image_depth}
    \begin{tabular}{l@{\hspace{0.5em}}c cc@{\hspace{1em}}cc@{\hspace{1em}}cc}
        \toprule
        \multirow{2}{*}{\textbf{Method}} & \multirow{2}{*}{\textbf{Ensemble}} 
            & \multicolumn{2}{c}{\textbf{DIODE~\cite{vasiljevic2019diode}}} 
            & \multicolumn{2}{c}{\textbf{KITTI~\cite{geiger2012we}}} 
            & \multicolumn{2}{c}{\textbf{ETH3D~\cite{schops2017multi}}} \\
        \cmidrule(lr){3-4} \cmidrule(lr){5-6} \cmidrule(lr){7-8}
        &  & AbsRel$\downarrow$ & $\delta1\uparrow$ 
            & AbsRel$\downarrow$ & $\delta1\uparrow$ 
            & AbsRel$\downarrow$ & $\delta1\uparrow$ \\
        \midrule
        Marigold$^\dagger$                & \xmark
           & {$31.0$} & {$77.2$} & $10.5$ & $90.4$ & {$7.1$} & {$95.1$} \\
        Marigold$^\dagger$                & \cmark  
            & \underline{$30.8$} & \underline{$77.3$} & \underline{$9.9$} & \underline{$91.6$} & {$\mathbf{6.5}$} & $\mathbf{96.0}$ \\
        Marigold$^\dagger$ w/ {\ourmodel} & \cmark 
            & $\mathbf{30.2}$ & $\mathbf{77.8}$ & $\mathbf{9.8}$ & $\mathbf{91.7}$ & $\underline{6.4}$ & \underline{$95.9$} \\
        \bottomrule
    \end{tabular}
    \vspace{-1em}
\end{table}

 \begin{figure}[!t]
    \centering
    \includegraphics[width=1\linewidth]{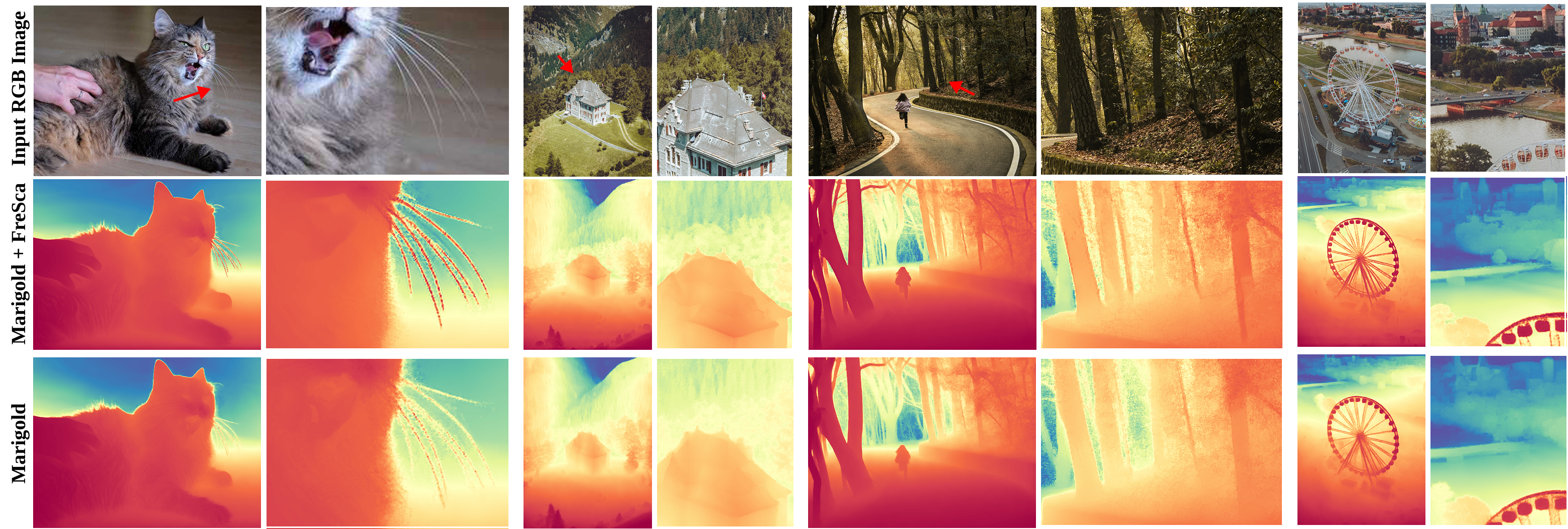}
    \caption{\textbf{\ourmodel sharpens depth predictions.} From top to bottom: input RGB, Marigold + \ourmodel, and Marigold. Red arrows highlight where our method recovers clearer shapes and reduces blur.}
    \vspace{-0.5em}
    \label{fig:image_depth}
\end{figure}

\subsection{Task: Monocular Depth Estimation}
\label{sec:image_depth}

Monocular depth estimation recovers 3D scene geometry from a single imag -- a key capability for autonomous driving, robotics, and augmented reality. Despite its intrinsic 2D to 3D ambiguity, latent diffusion methods like Marigold~\cite{ke2023repurposing}, which fine‐tunes only the denoising U-Net of Stable Diffusion~\cite{stablediffusion} on synthetic RGB-D data, achieve strong zero‐shot performance on real‐world benchmarks without ever using real depth maps. 

 While it generalizes well, it can miss fine details and misestimate distant objects. To address this, we equip Marigold with {\ourmodel} boosting its high-frequency noise components ($h>1$, $l=1$) while leaving the low frequencies intact. Specifically, Marigold’s predictor $\epsilon_t = \epsilon_\theta(\mathbf{d}_t, \mathrm{x}, t)$~\cite{ke2023repurposing} runs with fixed classifier-free guidance ($\omega=1$) and relies solely on the conditional branch. Therefore, we apply \texttt{\ourmodel} directly to the predicted noise $\epsilon_t$, since this noise encodes the semantic information necessary for accurate, detail-rich depth estimation.

As \cref{table:image_depth} shows, integrating {\ourmodel} consistently outperforms Marigold baselines (with or without ensemble) on DIODE~\cite{vasiljevic2019diode}, KITTI~\cite{geiger2012we}, and ETH3D~\cite{schops2017multi}, achieving leading AbsRel and $\delta_1$ metrics. Unlike ensembling, which can oversmooth, our frequency‐based adjustment yields more deterministic, accurate depth maps, recovering fine structures and sharp edges (see \cref{fig:image_depth}).

\begin{table}[t]
    \centering
    \caption{Image editing results evaluated by both generative metrics (FID-30k and CLIP-text) and human-centric VLM metrics (Success Rate and Quality).}
    \label{tab:combined_editing}
    \footnotesize
    \begin{tabularx}{\textwidth}{lXcccc}
        \toprule
        & & FID-30k $\downarrow$ & CLIP-text (\%) $\uparrow$ & Success Rate (\%) $\uparrow$ & Quality \\
        \midrule
        Edited-Friendly DDPM~\cite{huberman2024edit}
            & & $255.5$      & $31.35$      & $75.0$       & $\mathbf{4.23}$ \\
        DDPM~\cite{huberman2024edit} w/ {\ourmodel}
            & & $\mathbf{253.4}$ & $\mathbf{31.54}$ & $\mathbf{80.0}$ & $4.18$ \\
        \midrule
        LEdits++~\cite{brack2024ledits++}
            & & $255.3$      & $31.08$      & $72.5$       & $4.08$ \\
        LEdits++~\cite{brack2024ledits++} w/ {\ourmodel}
            & & $\mathbf{255.0}$ & $\mathbf{31.34}$ & $\mathbf{72.5}$ & $\mathbf{4.18}$ \\
        \bottomrule
    \end{tabularx}
    \vspace{-1em}
\end{table}
\begin{figure}[t]
    \centering
    \includegraphics[width=1\linewidth]{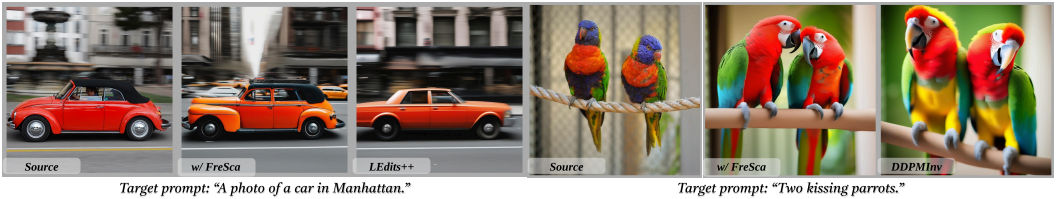}
    \caption{Editing results from LEdits++~\cite{brack2024ledits++} and DDPM inversion~\cite{huberman2024edit} with or without \ourmodel.}
    \label{fig:editing}
    \vspace{-1em}
\end{figure}
\subsection{Task: Text-guided Image Editing}
\noindent \textbf{Dataset and Baselines.} We conduct experiments on the public image editing dataset TEdBench~\cite{kawar2023imagic}, which comprises 40 images from diverse categories paired with various editing prompts. 
\ourmodel can be seamlessly integrated into existing image editing frameworks without altering their core architectures. Accordingly, we benchmark our approach against training-free methods, including LEdits++\cite{brack2024ledits++} and Edited-Friendly DDPM Inversion\cite{huberman2024edit}, strictly following their prescribed settings.

\noindent\textbf{Evaluation Protocol.}
For quantitative comparison, we fix the CFG $\omega=15$ for all methods. In our framework, we set $l=1$ for both, while applying $h=1.2$ for Edited‐Friendly DDPM Inversion and $h=2.0$ for LEdits++, using a spatial cutoff radius of 20 in both cases. Further discussion on $l$ and $h$ choices can be found in the supp. We measure editing fidelity with the CLIP‐text similarity~\citep{clip} against the target prompt, and assess overall image quality via FID‐30k~\citep{fid}. Additionally, we perform qualitative evaluation using the large vision–language model InternVL2.5‐8B~\citep{chen2024expanding}.

\noindent\textbf{Results.}
As reported in \cref{tab:combined_editing}, integrating {\ourmodel} into both Edited‐Friendly DDPM Inversion and LEdits++ consistently boosts CLIP‐text scores and reduces FID, demonstrating that selective amplification of high‐frequency detail strengthens the target edit, preserves image fidelity, and increases the editing success rate. Qualitative examples in \cref{fig:editing} further illustrate these enhancements.

\begin{figure}[h]
    \centering
    \includegraphics[width=0.99\linewidth]{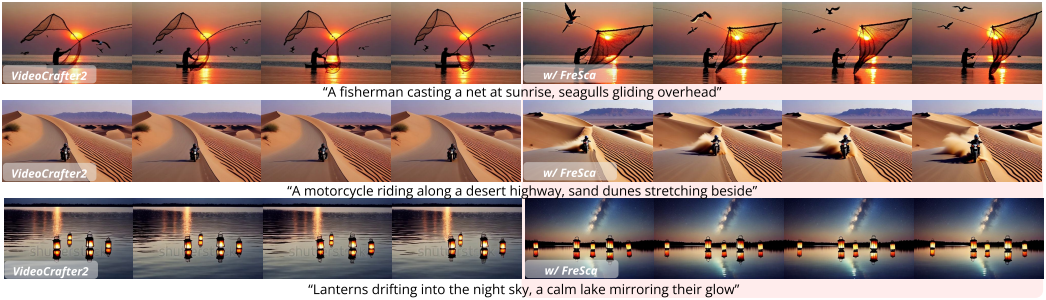}
    \caption{\ourmodel enhances VideoCrafter2’s~\cite{chen2024videocrafter2} video generation quality at no additional cost.}
    \label{fig:videocrafter2}
\vspace{-1em}
\end{figure}

\subsection{Task: Text to Video Generation}
\ourmodel's applicability is not limited to static image tasks; we demonstrate its effectiveness in the dynamic domain of video generation. We integrate \ourmodel into VideoCrafter2~\cite{chen2024videocrafter2}, an open-source video diffusion model. By modulating solely the high-frequency components of the predicted noise, we achieve improvements in video quality and fidelity without any model retraining. As illustrated in \cref{fig:teaser,fig:videocrafter2}, \ourmodel enhances motion coherence, preserves intricate details, and mitigates text-video misalignment. This underscores \ourmodel's significant potential and versatility across diverse diffusion models.

\section{Conclusion}
This paper introduced {\ourmodel}, a novel framework enabling fine-grained, disentangled control over latent diffusion models through frequency-domain manipulation. By targeting the semantically rich classifier-free guidance noise difference $\Delta\epsilon_t$, {\ourmodel} decomposes it into frequency bands, applying scaled adjustments with dynamic cutoffs. This model-agnostic, plug-and-play approach is shown to effectively control visual attributes across various models (SDXL, SD3) and tasks (image generation, editing, depth estimation, video synthesis). {\ourmodel} not only provides practical creative control but also contributes to understanding frequency components in LDMs. Future work could explore advanced spectral techniques and learned control strategies.
 
\bibliographystyle{unsrtnat}
\bibliography{main}

\newpage
\appendix

\section{Analysis on Text to Image Generation}

\subsection{Analysis of Frequency Scaling Parameters and Cutoff Strategies}
\noindent\textbf{Effects of Frequency Scaling Factors $l, h$, and Cutoff Ratio $r_0$.} We investigate the impact of our low-frequency scaling factor $l$, high-frequency scaling factor $h$, and the cutoff ratio $r_0$ under two distinct frequency cutoff strategies. 

\begin{figure}[t]
 \centering
 \includegraphics[width=1\linewidth]{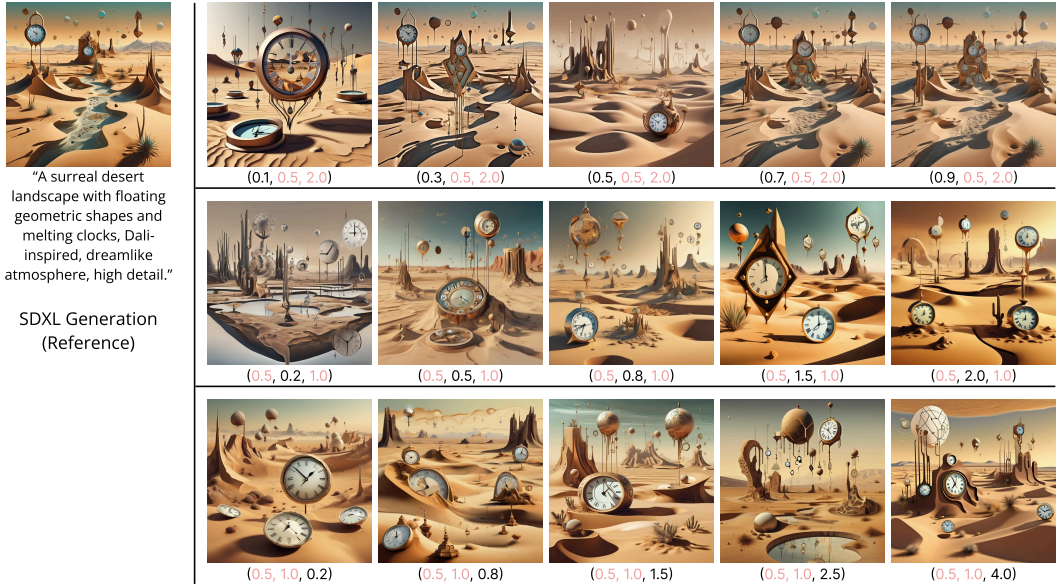}
 \caption{Visual effects of varying cutoff thresholds and scaling factors ($r_0$, $l, h$) using the Spatial-Ratio Cutoff strategy.} %
 \label{fig:hyper_spatio}
\end{figure}

\noindent \textbf{Spatial-Ratio Cutoff Strategy.} This strategy defines the cutoff frequency based on a spatial frequency ratio $r_0$, where a fixed proportion of the lowest spatial frequencies are low-frequency components (\cref{fig:hyper_spatio}).
\begin{itemize}[leftmargin=1.5em]
 \item \textbf{Impact of Cutoff Ratio $r_0$:} Low $r_0$ ($0.1$) results in most frequencies being treated as high-frequency, leading to strong detail amplification and potential noise with high $h$. Increasing $r_0$ towards $0.3 \sim 0.5$ designates a larger portion as low-frequency, yielding a more balanced mix of structure and detail enhancement. High $r_0$ means most frequencies are low-frequency, resulting in smoother images with subtle detail ``pop'' even with high $h$, as fewer high-frequency components exist.
 \item \textbf{Impact of Low-Frequency Scaling Factor $l$:} Varying $l$ scales coarse structures (fixed $r_0, h=1.0$). Low $l$ ($0.2$) heavily suppresses coarse forms, emphasizing edges and textures. $l$ values $ 0.5\sim0.8$ attenuate coarse structures to a lesser degree, balancing form and detail. $l \ge 1.5$ enhances coarse structures, potentially overpowering fine details.
\item \textbf{Impact of High-Frequency Scaling Factor $h$:} Varying $h$ scales fine details and textures (fixed $r_0, l$). Low $h$ ($ 0.2$) suppresses details, making the image less sharp. $h = 1.5$ provides noticeable sharpening without significant artifacts. Very high $h$ ($= 4$) causes strong, often artifact-prone sharpening, potentially useful for stylized effects but detrimental to realism.
\end{itemize}

\begin{figure}[t]
 \centering
 \includegraphics[width=1\linewidth]{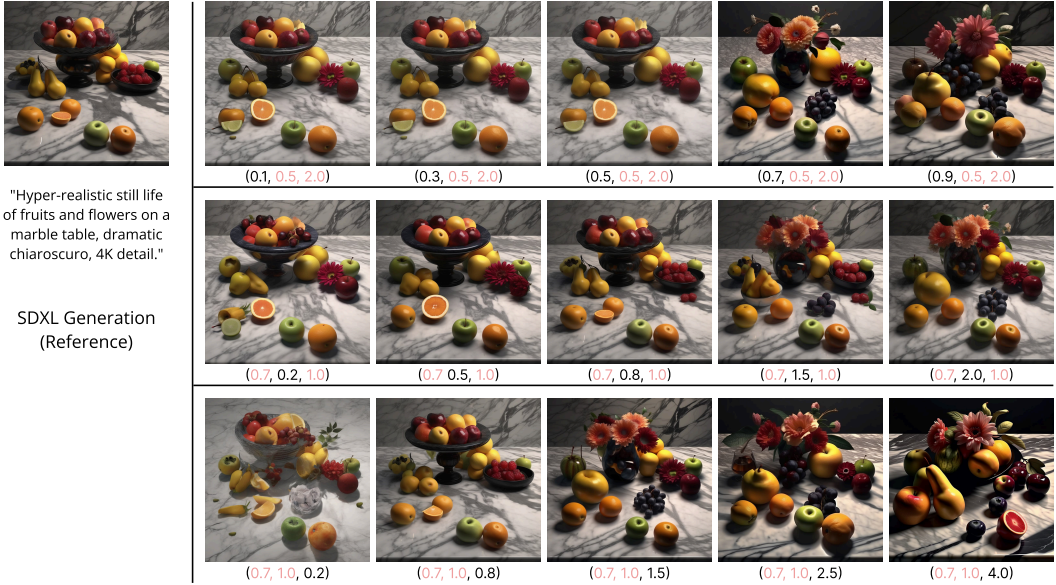}
 \caption{Visual effects of varying cutoff thresholds and scaling factors ($r_0$, $l, h$) using the Energy-based Cutoff strategy.} %
 \label{fig:hyper_energy}
\end{figure}

\noindent\textbf{Energy-based Cutoff Strategy.} This strategy defines the cutoff frequency based on the cumulative energy spectrum, with $r_0$ as the energy threshold (\cref{fig:hyper_energy}).
\begin{itemize}[leftmargin=1.5em]
    \item \textbf{Impact of Cutoff Ratio $r_0$:} Varying $r_0$ redistributes spectral energy between the low and high-frequency bands. For low $r_0$ (e.g., $0.1 \sim 0.5$), most energy is in the high-frequency band; scaling the low-frequency components has minimal impact, highlighting the energy distribution. For high $r_0$ ($ 0.5 \sim 0.9$), most energy is low-frequency. In this case, scaling factors become more influential, particularly a high $h$ which enhances finer details within the remaining high frequencies. The sensitivity of local structures to low $h$ (e.g., 0.2) further demonstrates the crucial role of high frequencies.
    \item \textbf{Impact of High-Frequency Scaling ($h$):} (e.g., $r_0=0.7$, $l=1.0$) Increasing $h$ amplifies fine details. $h=1.0$ is baseline. $h=1.5-2.0$ yields mild to clear detail enhancement without significant artifacts. $h \ge 2.5$ leads to strong sharpening and potential artifacts. Photographic realism is best achieved with a moderate $h \in [1.5, 2.5]$; $h > 3.0$ suits stylized effects.
    \item \textbf{Impact of Low-Frequency Scaling ($l$):} (e.g., $r_0=0.7$, $h=1.0$) Varying $l$ scales coarse structures. $l<1$ can negatively impact prompt adherence, while $l>1$ appears to improve it. Changes to local structure and style from varying $l$ are generally more subtle than those from $h$.
\end{itemize}

\noindent\textbf{Summary.} Optimal parameter selection balances structural preservation and frequency component scaling. The Energy-based cutoff strategy offers good interpretability, and the significant role of high frequencies allows for diverse applications.

\subsection{Fine-grained Changes with Different Cutoff Thresholds for Pixel \& VAE Spaces}
\begin{figure}[!h]
    \centering
    \includegraphics[width=0.7\linewidth]{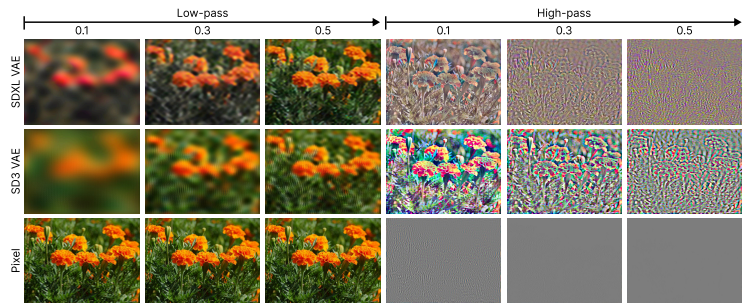}
    \caption{Frequency decomposition with different cut-off thresholds (a more fine-grained version of \cref{fig:vae_pixel}. }
    \label{fig:cutoff}
\end{figure}

To complement \cref{fig:vae_pixel}, we provide a more detailed change as visualized in \cref{fig:cutoff}.

\subsection{Quantative and Visual Effects of Frequency Scaling Parameters ($h$, $l$, $r_0$)}
We quantitatively analyze the effects of varying the hyperparameters $h$, $l$, and $r_0$ using the energy-based cutoff strategy) on the SD3 base model, as presented in \cref{tab:ablation_sd3}.

Our observations reveal that enhancing high-frequency components effectively improves image-text alignment (higher CLIP score), though it slightly degrades the generation FID. Conversely, enhancing low-frequency components yields inverse effects: a lower (better) FID but a diminished (worse) CLIP score.

\begin{figure}[t]
    \centering
    \includegraphics[width=0.8\linewidth]{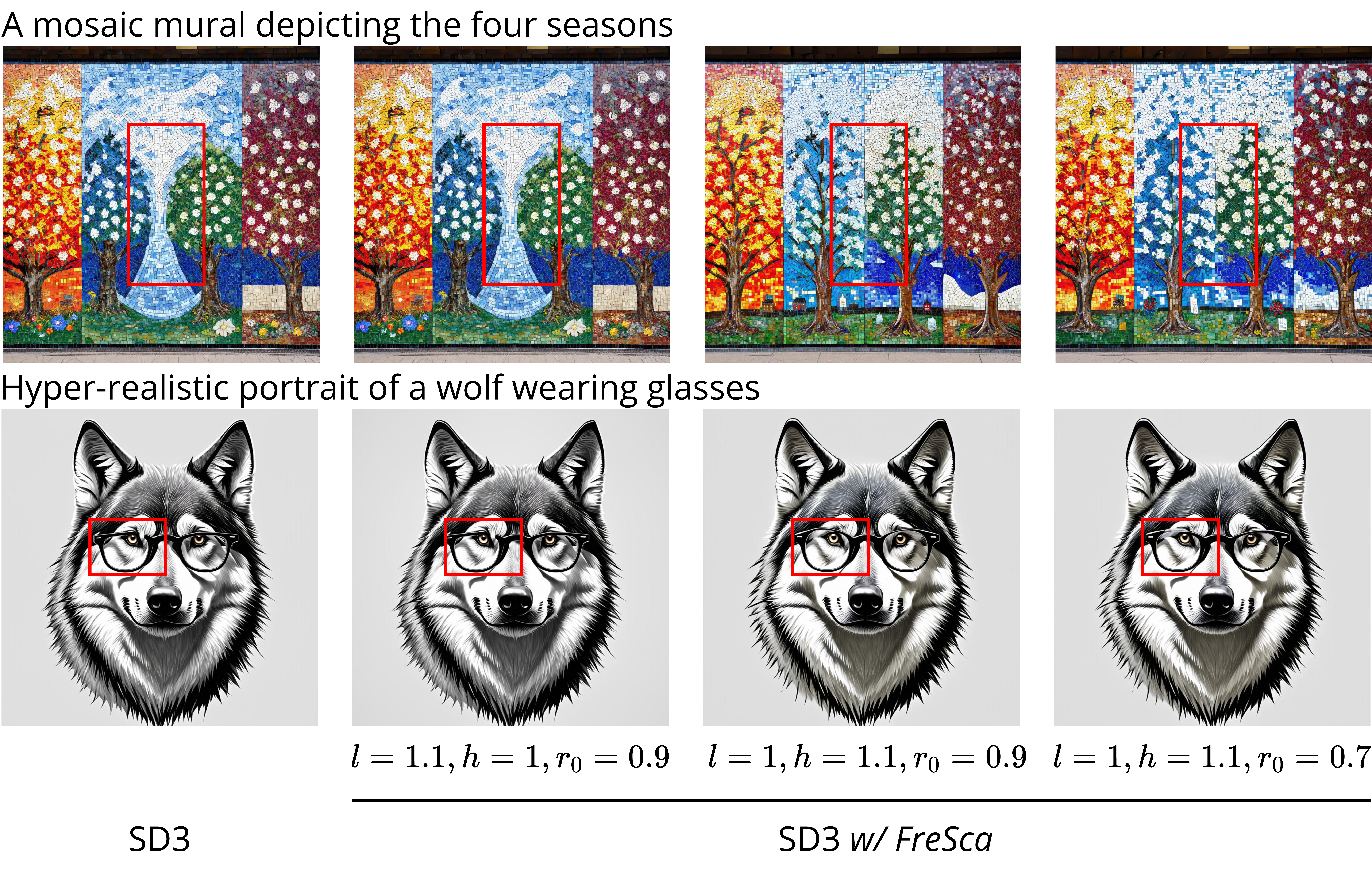}
    \caption{Visualization of \ourmodel applied to SD3, showing the effect of varying the hyperparameters $h$, $l$, and $r_0$. Red box shows the region of interest, where increasing high freqeuncy bring higher image-prompt alignment compared to the baseline, while improving low-frequency marginally improve the generation FID.}
    \label{fig:ablation_sd3}
\end{figure}

As visually demonstrated in \cref{fig:ablation_sd3}, the enhancement of high-frequency components is crucial as it significantly improves prompt alignment and facilitates better instruction following. Importantly, the minor quantitative degradation in FID (e.g., from 219.96 to 220.47) does not noticeably impact the subjective quality of the generated images.

Therefore, frequency scaling proves to not only be a useful technique for controlling image characteristics, but also in affecting diffusion-based image representations. Optimizing the combination of different hyperparameter sets for specific generation goals is an important direction for future work.

\begin{table}[t]
    \centering
    \caption{Ablation study of \ourmodel applied to SD3, showing the effect of slightly varying the hyperparameters $h$, $l$, and $r_0$ on the two generation evaluation metrics FID and CLIP-text scores.}
    \begin{tabular}{lcccccc}
        \toprule
        \textbf{Method} & \textbf{$h$} & \textbf{$l$} & \textbf{$r_0$} & \textbf{FID} $\downarrow$ & \textbf{CLIP Score (\%)} $\uparrow$ \\
        \midrule
        \rowcolor{gray!15}
        SD3 (baseline) & -- & -- & -- & 219.96 & 16.24 \\
        \addlinespace[0.5ex]
        \multicolumn{6}{l}{\textit{SD3 + \ourmodel}} \\
        \quad w/ \ourmodel & 1.0 & 1.1 & 0.9 & 219.70 $\textcolor{green}{\blacktriangledown}$ & 16.23 $\textcolor{red}{\blacktriangledown}$ \\
        \quad w/ \ourmodel & 1.1 & 1.0 & 0.9 & 220.47 $\textcolor{red}{\blacktriangle}$& 16.25 $\textcolor{green}{\blacktriangle}$\\
        \quad w/ \ourmodel & 1.1 & 1.0 & 0.7 & 220.57 $\textcolor{red}{\blacktriangle}$ & 16.30 $\textcolor{green}{\blacktriangle}$\\
        \quad w/ \ourmodel & 1.1 & 1.0 & 0.5 & 219.98 $\textcolor{red}{\blacktriangle}$ & 16.23 $\textcolor{red}{\blacktriangledown}$\\
        \bottomrule
    \end{tabular}
    \label{tab:ablation_sd3}
\end{table}

\subsection{Understanding Step-wise Dynamics of High-Frequency Scaling}

\begin{table}[t]
 \centering
\caption{Ablation study of \ourmodel applied to SD3, showing the effect of varying the high-frequency scaling schedule on FID and CLIP-text scores.}
 \begin{tabular}{lcccccc}
 \toprule
\textbf{Method} & \textbf{$h$} & \textbf{$l$} & \textbf{$r_0$} & \textbf{FID} $\downarrow$ & \textbf{CLIP Score (\%)} $\uparrow$ \\
 \midrule
 \quad w/ \ourmodel & 1.1 & 1.0 & 0.9 & 220.47 & 16.25 \\
 \quad w/ \ourmodel Linear Growth & 1.1 & 1.0 & 0.9 & 220.60 $\textcolor{red}{\blacktriangle}$ & 16.19 $\textcolor{red}{\blacktriangledown}$\\
\quad w/ \ourmodel Linear Decay & 1.1 & 1.0 & 0.9 & 219.69 $\textcolor{green}{\blacktriangledown}$ & 16.23 $\textcolor{red}{\blacktriangledown}$\\
 \bottomrule
 \end{tabular}
 \label{tab:schedule}
\end{table}
\begin{figure}[h] %
\centering
 \includegraphics[width=0.8\linewidth]{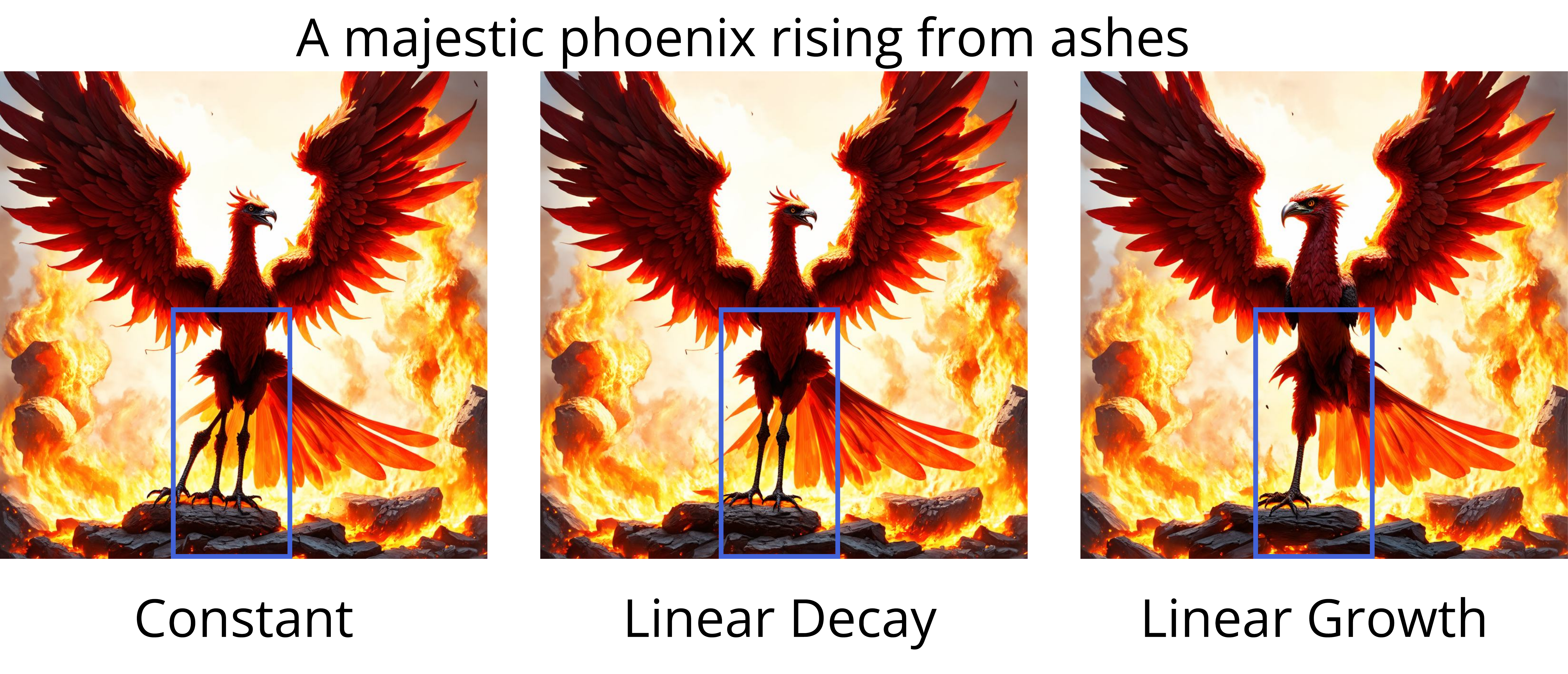}
 \caption{Visual results comparing different high-frequency scaling schedules.} %
 \label{fig:h_schedule}
\end{figure}

\begin{figure}[h!] %
\centering
\includegraphics[width=0.7\linewidth]{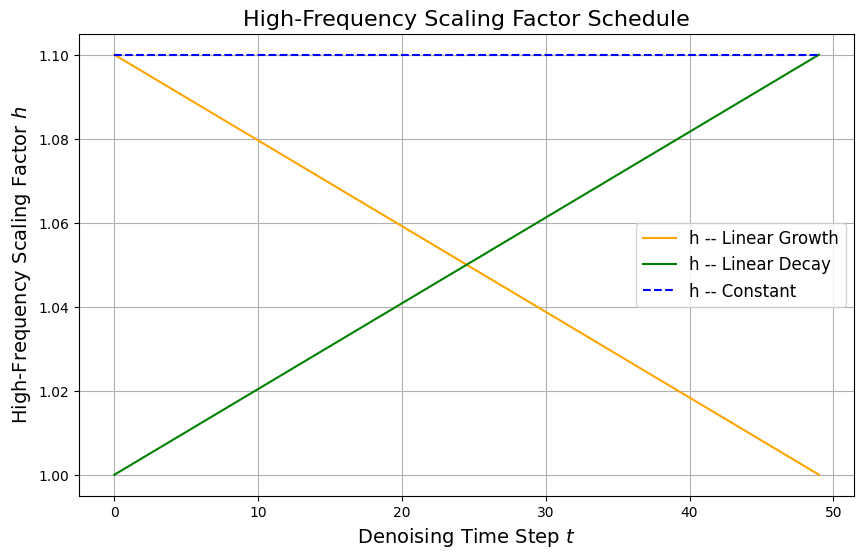}
 \caption{Illustration of Linear Decay and Linear Growth schedules for the high-frequency scaling factor $h$ over denoising steps $t$.} %
 \label{fig:schedule}
\end{figure}

Beyond the observation that enhancing high-frequency components improves image-prompt alignment, we explore whether the scaling factor $h$ benefits from a time-dependent schedule. As illustrated in \cref{fig:step}, the high-frequency components of $\Delta\epsilon$ intensify as the denoising process progresses.

To investigate the importance of this dynamic, we introduce two scheduling strategies for $h$, defined over the total 50 denoising steps ($t \in [0, 49]$), as shown in \cref{fig:schedule}:

\textbf{Linear Decay:} $h(t) = \frac{49 - t}{49} \cdot (h_{max} - 1) + 1$
\quad \textbf{Linear Growth:} $h(t) = h_{max} - \frac{49 - t}{49} \cdot (h_{max} - 1)$

where $h_{max}$ is the maximum high-frequency scaling factor (e.g., 1.1 in our ablation).

As detailed in \cref{tab:schedule} (using $h_{max}=1.1$), adopting a \textbf{Linear Decay} strategy for $h$ yields better FID while slightly reducing the CLIP score compared to a constant $h$. This suggests that attenuating high-frequency factors in earlier steps provides better image preservation, as the higher magnitude of high-frequency components in later steps makes their scaling more impactful. Conversely, the \textbf{Linear Growth} strategy did not contribute positively to either metric.

As verified qualitatively in \cref{fig:h_schedule}, dynamic adjustment of high-frequency components can lead to more faithful results. This time-aware scheduling serves as an optional strategy for practical implementation, warranting further investigation in future work.

\subsection{More Image Generation Results}
We show more generation results from SDXL and SD3 with or without \ourmodel below.
\begin{figure}[h!]
    \centering
    \includegraphics[width=0.82\linewidth]{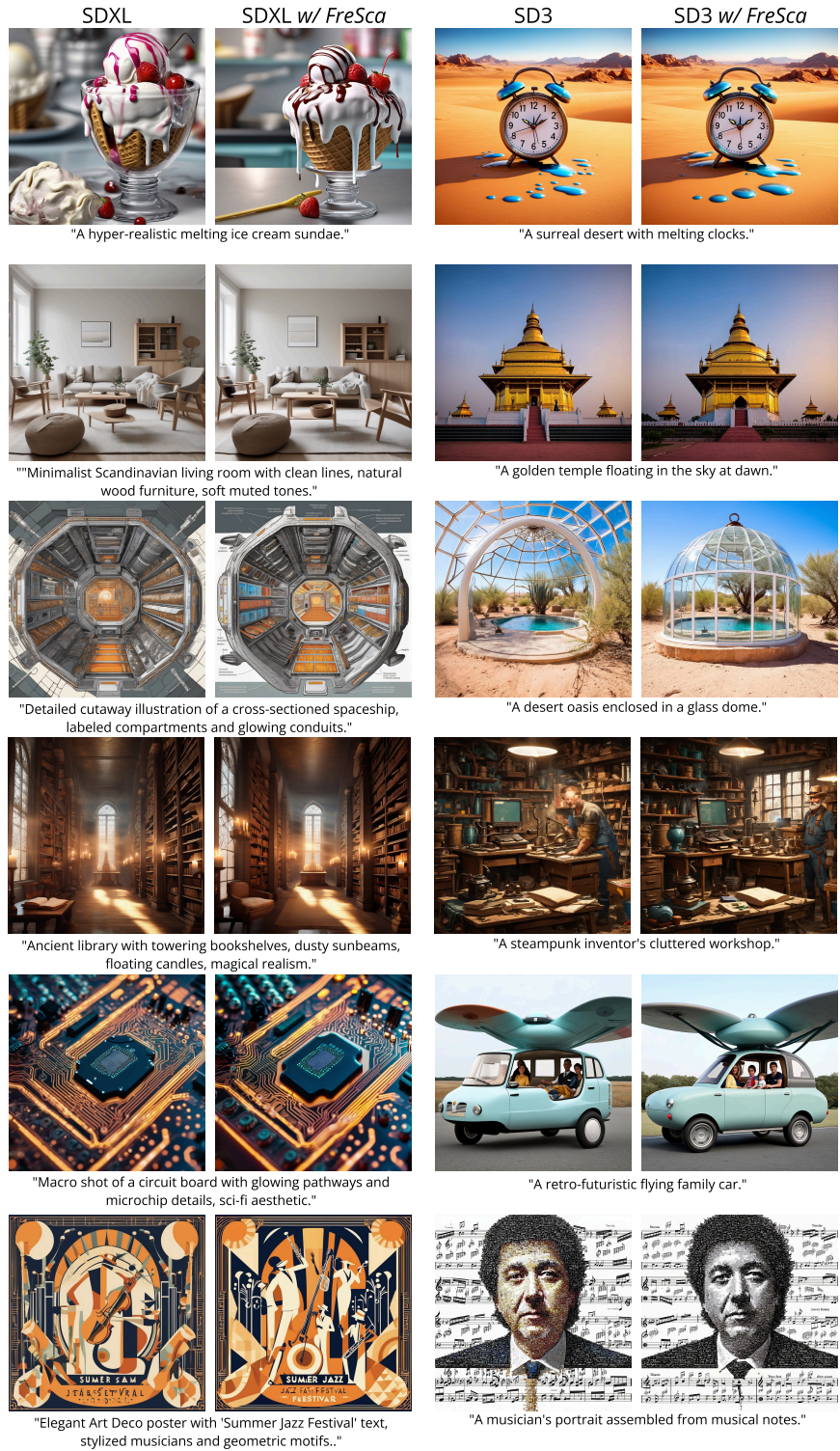}
    \caption{More generation examples from SDXL and SD3 with or without \ourmodel.}
    \label{fig:more_vis}
\end{figure}
\section{Results on Text to Video Generation}

We have created a project page to illustrate our method and showcase our results. \textbf{We strongly encourage readers to visit this webpage}.
 
\section{More Details on the Editing Task}

\noindent\textbf{Details about TEdBench~\cite{kawar2023imagic} Dataset}
The complete list of image names and their target text prompt mappings we used for evaluation are shown in \cref{tab:image_target_text}.
\begin{table}[h]
    \centering
    \begin{tabular}{|c|l|}
        \hline
        \textbf{Image Name} & \textbf{Target Text} \\
        \hline
        dog2\_standing.png & A photo of a sitting dog. \\
        tennis\_ball.jpeg & A photo of a tomato in a blue tennis court. \\
        zebra.jpeg & A photo of a horse. \\
        red\_car.jpeg & A photo of a car in Manhattan. \\
        bird.jpeg & A photo of a bird spreading wings. \\
        box.jpeg & A photo of an open box. \\
        cat.jpeg & A photo of a cat wearing a hat. \\
        cat\_3.jpeg & A photo of a cat wearing a hat. \\
        dog\_with\_shirt.jpg & A dog smoking a cigar. \\
        dog\_01.jpeg & A photo of a sitting dog. \\
        vase\_01.jpeg & A photo of a vase of red roses. \\
        door.jpeg & A photo of an open door. \\
        couple\_beach.jpeg & A photo of a couple holding hands on a beach. \\
        open\_book.jpeg & A photo of a closed book. \\
        empty\_street.jpeg & A busy congested street. \\
        black\_shirt.jpeg & A person with crossed arms. \\
        bear3.jpeg & A black bear walking in the grass next to red flowers. \\
        milk\_cookie.jpeg & A cookie next to a glass of juice. \\
        chibi.jpeg & Image of a cat wearing a floral shirt. \\
        giraffe.jpeg & A giraffe with a short neck. \\
        apples.jpeg & A basket of oranges. \\
        new\_cat\_3.jpeg & A photo of a sleeping cat. \\
        chair\_1.jpeg & A knocked down chair. \\
        flamingo.jpeg & A sitting flamingo. \\
        banana\_1.jpeg & A photo of a sliced banana. \\
        cake\_1.jpeg & A photo of a birthday cake. \\
        tree\_1.jpeg & A photo of a dead tree. \\
        teddy\_1.jpeg & A photo of a teddy bear doing pushups. \\
        white\_horse1.png & A white horse in a grass field. \\
        white\_horse2.png & A jumping horse. \\
        prague.png & A cyclist riding in a street. \\
        bird.png & A bird looking backwards. \\
        goat\_and\_cat.jpg & A goat and a cat hugging. \\
        elephant.jpeg & A person riding on an elephant. \\
        road1.png & An image of a post-apocalyptic road. \\
        egg\_tree.jpeg & A cracked egg. \\
        two\_dogs\_with\_checkered\_shirts1.jpg & Two dogs growling at each other. \\
        pizza1.png & Pizza with pepperoni. \\
        drinking\_horse.png & A horse raising its head. \\
        bird-g83440b9c4\_1920.jpg & Two kissing parrots. \\
        \hline
    \end{tabular}
    \caption{Image names and their corresponding target texts.}
    \label{tab:image_target_text}
\end{table}

\begin{figure}[!t]
    \centering
    \includegraphics[width=0.6\linewidth]{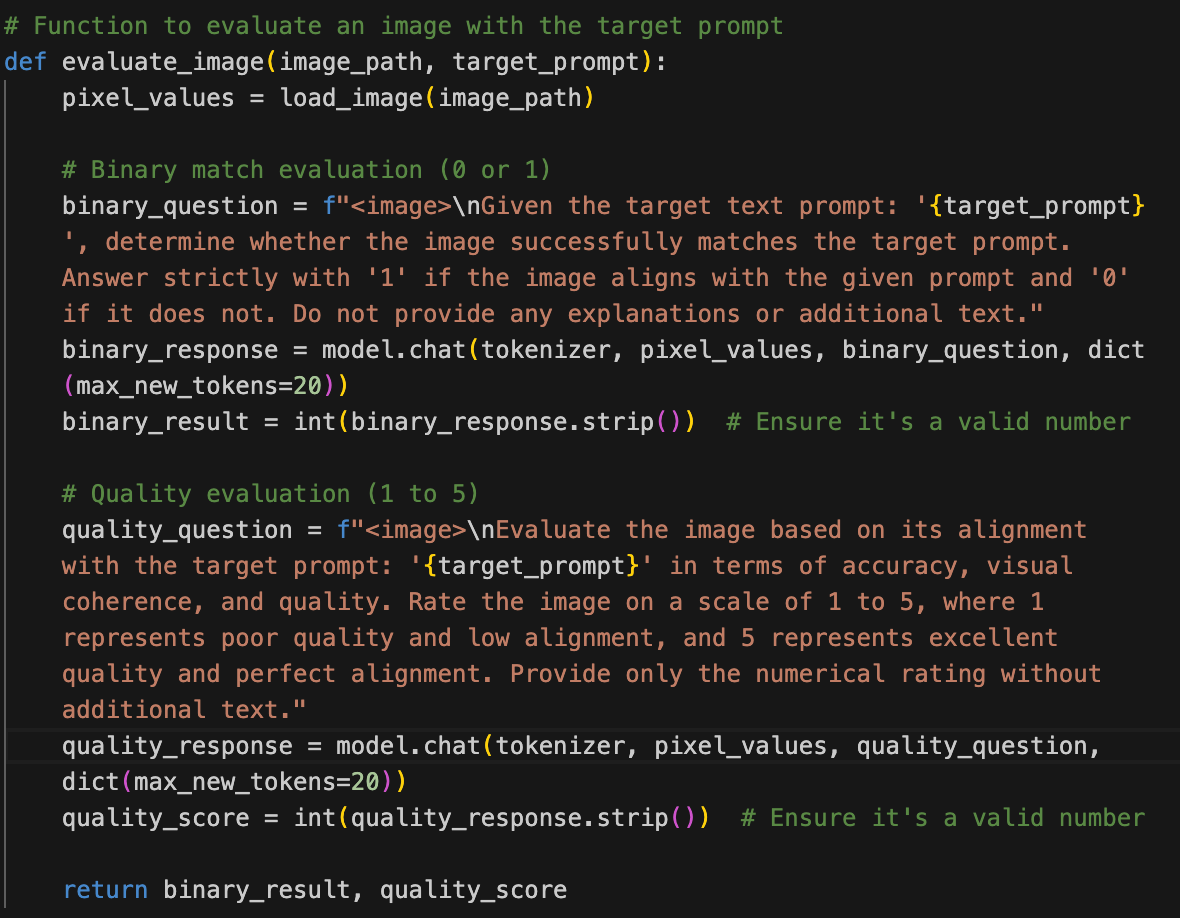}
    \caption{Prompts designed for LVLM evaluation.}
    \label{fig:LVLM}
\end{figure}

\noindent \textbf{Success Rate \& Quality Metic in \cref{tab:combined_editing}.} We further evaluate the edited images using the large vision-language model InternVL2.5-8B~\cite{chen2024expanding}. This model provides a binary decision (0 or 1) to indicate whether the editing was successful and assigns a qualitative score on a scale of 1 to 5—where 1 denotes poor quality and 5 reflects excellent performance in both concept fidelity and overall image quality. As shown in \cref{tab:combined_editing}, incorporating \ourmodel not only improves the overall quality of the edited outputs but also increases the editing success rate. This demonstrates the effectiveness of our approach in achieving high-quality, semantically faithful edits. The prompt design for obtaining these metrics are shown in \cref{fig:LVLM}.

\begin{figure}[!h]
    \centering
    \includegraphics[width=0.6\linewidth]{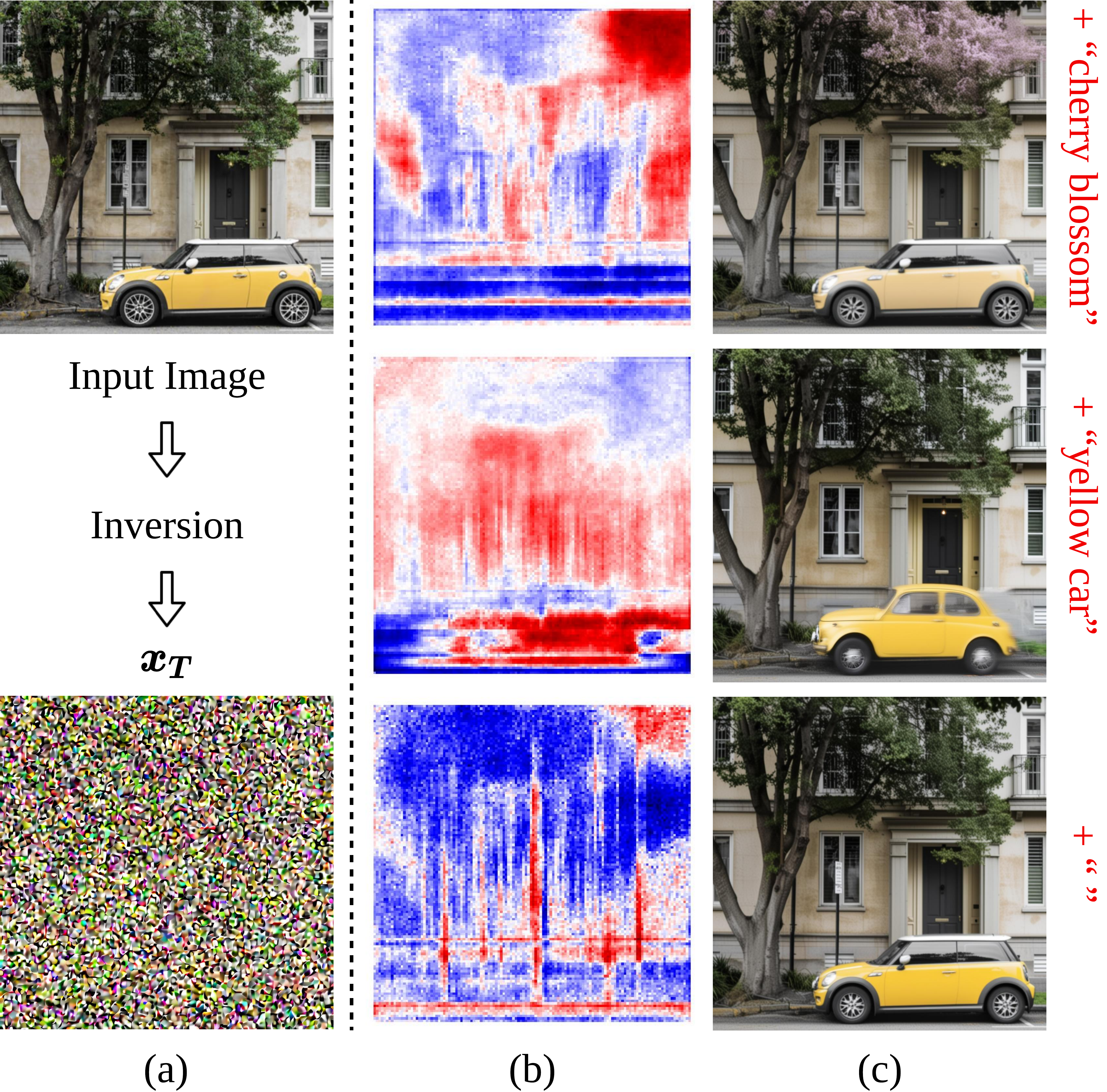}
    \caption{\textbf{Key components for image editing:} (a) latent vector $\mathbf{x}_T$ is obtained through inversion techniques; (b) the visualizations of $\Delta\epsilon$ at the first editing step; (c) the final edited output.}
    \label{fig:paradigm}
\end{figure}

\noindent\textbf{Role of $\Delta\epsilon$ in Image Editing}
Intuitively, the noise difference term $\Delta\epsilon_t$ in image editing encodes spatial regions corresponding to the target prompt. We validate this through three distinct editing scenarios depicted in \cref{fig:paradigm}: replacement editing, self-editing, and using an unrelated prompt. As illustrated, the $\Delta\epsilon_t$ maps clearly show activation in prompt-relevant areas for semantically related prompts, while exhibiting diffuse or random patterns for unrelated ones. This analysis leads to three key observations about the editing process:
\noindent\textbf{1. Semantically Rich Inversion:} The inverted initial latent $\mathbf{x}_T$ preserves essential input semantics, aligning with the target prompt $\mathbf{c}'$.
\noindent\textbf{2. $\Delta\epsilon_t$ As Prompt Proxy:} The noise prediction difference $\Delta\epsilon_{t}$ effectively isolates and spatially localizes the representation of the target prompt.
\noindent\textbf{3. $\omega$ Modulates Edit Strength and Direction:} Given that $\Delta\epsilon_{t}$ represents the target concept, the scalar factor $\omega$ directly modulates the strength and determines the enhancement or suppression direction of the edit.

\begin{figure}[t]
    \centering
    \includegraphics[width=0.99\linewidth]{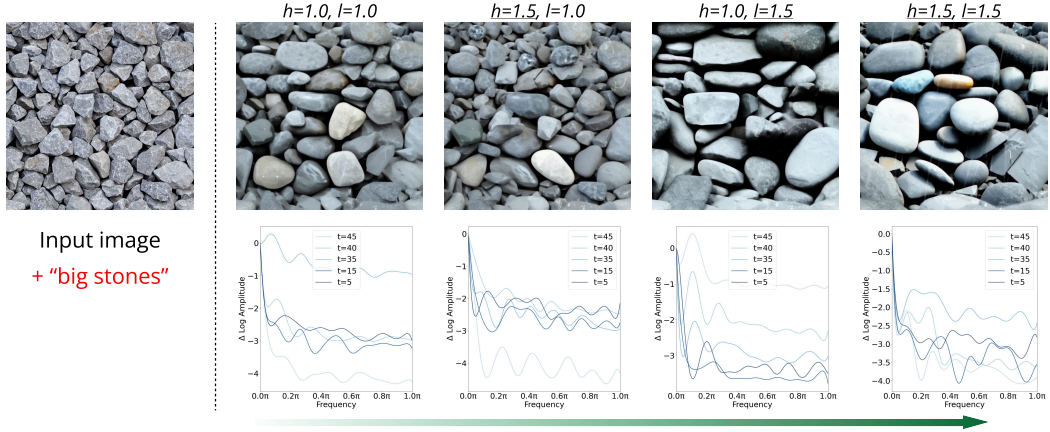}
    \caption{\textbf{Frequency scaling effects on the image editing task}:  We set the target prompt to increasing the size of stones and apply three different scaling strategies in the frequency domain: uniform scaling ($h=l=1.5$), low-frequency scaling ($l=1.5$, $h=1$), and high-frequency scaling ($h=1.5$, $l=1$). Each approach produces distinct effects.}
    \label{fig:editing_scaling}
\end{figure}

\subsection{Understanding Editing Dynamics via Fourier Analysis}
Here, we further study the roles of $\omega$ and $\Delta\epsilon_t$ in image editing, understand how frequency scaling works for this task. We analyze $\Delta\epsilon_t$ in the Fourier domain, decomposing it into low ($\Delta\epsilon_t^l$) and high ($\Delta\epsilon_t^h$) frequency components using a spatial-ratio cutoff threshold (with $r_0 = 0.3$).
We question if low- and high-frequency dynamics are equivalent. By introducing independent scaling factors $l$ and $h$, we found distinct roles. \cref{fig:editing_scaling} shows that asynchronous scaling (e.g., $l=1.5, h=1$) primarily affects structure, while $h=1.5, l=1$ adds texture. Also, the relative Fourier log-amplitude patterns are different when choosing different combinations of $l$ and $h$. Therefore, it reveals that low and high frequencies are not always synchronous, suggesting a need for flexible scaling.

\begin{figure}[b]
    \centering
    \includegraphics[width=0.7\linewidth]{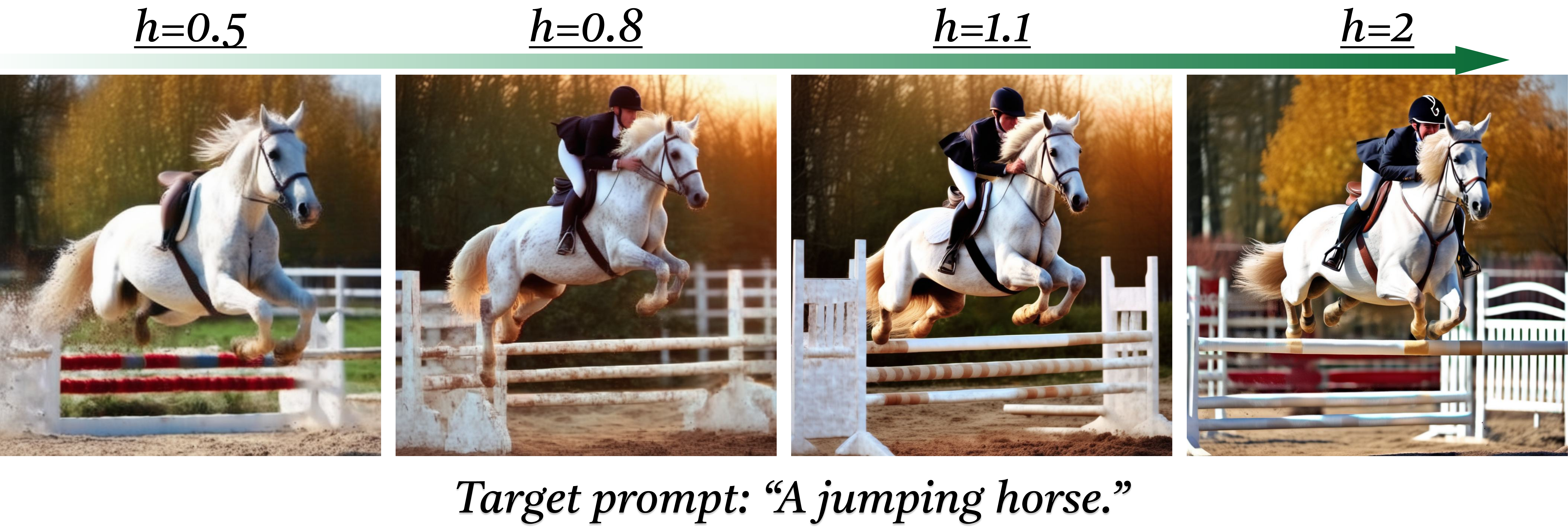}
    \caption{Continuous adjustment of high-frequency components. We scale the $h$ from $0.5$ to $2$ to examine its impacts on the editing performance.}
    \label{fig:trainsition}
\end{figure}

\begin{figure}[t]
    \centering
    \includegraphics[width=0.55\linewidth]{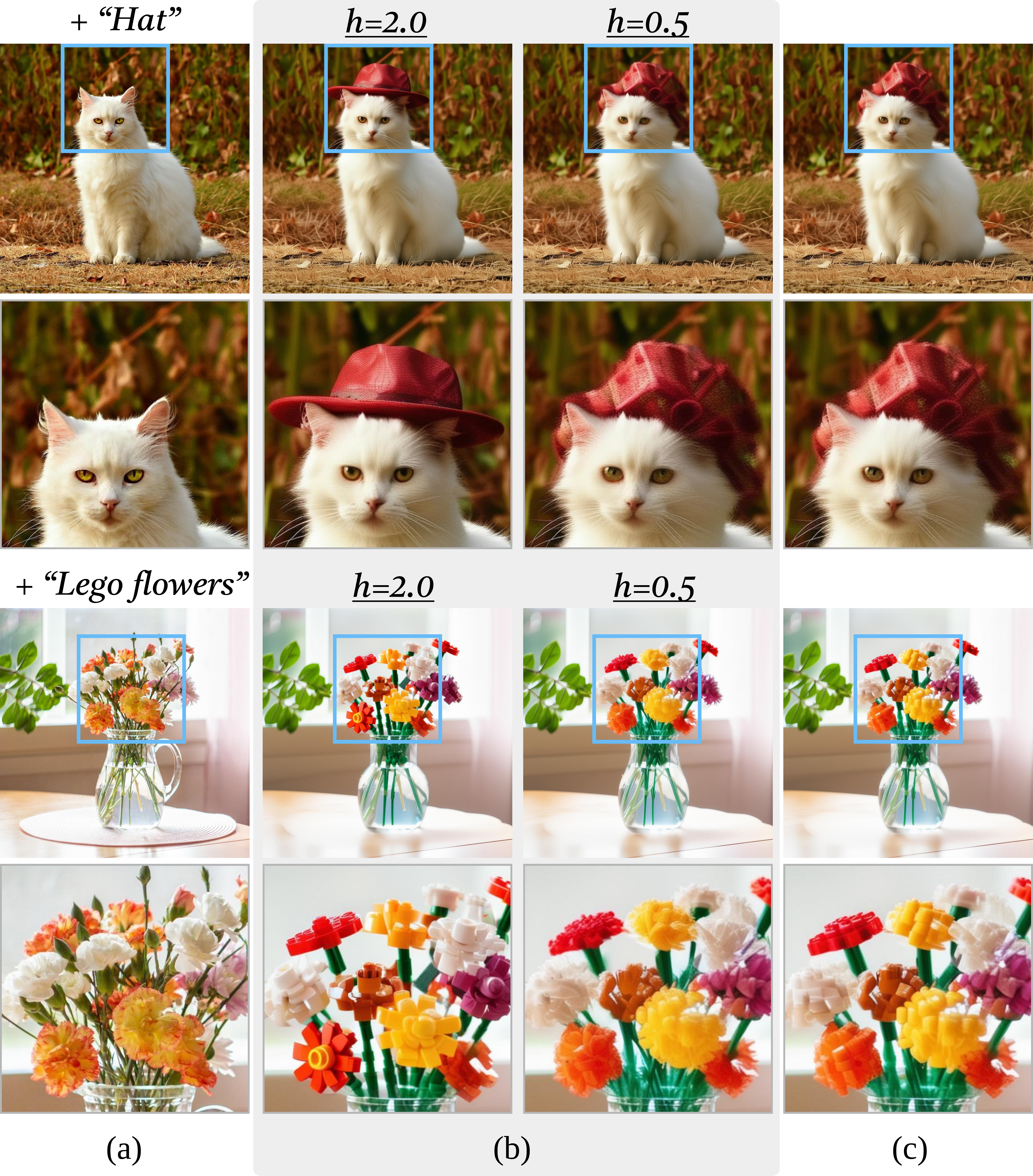}
    \caption{\textbf{Scaling up the high-frequency parts} ($h=2.0$) effectively enhances the editing fidelity.The red hat is successfully injected, and the edge of the LEGO flowers is sharpened. (a) input image, (b) results from \ourmodel with different $h$ being set, and (c) results from LEdits++. }
    \label{fig:freq_self}
\end{figure}

\noindent \textbf{Transition with varying $h$.} For a given image, altering $h$ from values below 1 to values above 1 produces an intriguing transition. When $h<1$, gradually increasing $h$ (e.g., from 0.5 to 0.8) introduces fundamental structural details, as evidenced by the appearance of the ``riding horse person.'' In contrast, when $h>1$, further increases enhance edges, contours, and other high-frequency features. These findings indicate that $h$ spans a scaling space that governs both high-frequency patterns and the underlying structural composition of the image, demonstrating that \ourmodel offers superior controllability compared to prior scaing space.

\noindent \textbf{The role of high-frequency scaling factors $h$.} As demonstrated \cref{fig:freq_self}, adjusting the high-frequency scaling factor $h$ produces two distinct effects: when $h>1$, the representation of shape, structure, and contour is enhanced, while setting $h<1$ introduces a counter-effect that pulls the edited result closer to the original image. This creates a practical trade-off between inducing more pronounced shape changes and better preserving the original structure. \ourmodel decouples these components, achieve varying levels of subtle control on $h$ without altering the primary editing direction.

\section{Experiment Configuration}
Here, we summarize the default configurations for getting results for different task in the main paper. Note that we do not massively search for the best combination of $h$, $l$, and $r_0$, but rather empircally pick a set for each task. Even without grid-search, \ourmodel works as an effective plug-and-play module for different models and different method. To observe, all task favors enhancing the high-frequency components while keeping its low-freq the same. In the following sections, we will show the effect of different roles for adjusting $h$, $l$, and etc.

\begin{table}[htbp]
  \centering
  \caption{Configuration settings for each task in the main paper.}
  \label{tab:configurations}
  \begin{tabular}{@{}l l c c c l l@{}}
    \toprule
    Task                                  & Baseline            & $h$  & $l$ & $r_0$ & Cutoff strategy           & Dataset   \\
    \midrule
    \addlinespace
    \multirow{2}{*}{Text-to-Image Generation}
                                          & SDXL~\cite{sdxl}                & 1.5  & 1   & 0.9   & \multirow{2}{*}{Energy-based} & \multirow{2}{*}{N/A}       \\
                                          & SD3~\cite{sd3}                 & 1.2  & 1   & 0.9   &                             &           \\
    \addlinespace
    \midrule
    \addlinespace
    \multirow{3}{*}{Monocular Depth Prediction}
                                          & \multirow{3}{*}{Marigold~\cite{ke2023repurposing}}
                                                               & 1.5  & 1   & 0.3   & \multirow{3}{*}{Spatial-ratio} & DIODE~\cite{vasiljevic2019diode}     \\
                                          &                     & 1.2  & 1   & 0.3   &                             & KITTI~\cite{geiger2012we}     \\
                                          &                     & 1.1  & 1   & 0.3   &                             & ETH3D~\cite{schops2017multi}     \\
    \addlinespace
    \midrule
    \addlinespace
    \multirow{2}{*}{Text-guided Image Editing}
                                          & LEDits++~\cite{brack2024ledits++}            & 2.0  & 1   & 0.3   & \multirow{2}{*}{Spatial-ratio} & \multirow{2}{*}{TEdBench~\cite{kawar2023imagic}}  \\
                                          & DDPM Inversion~\cite{huberman2024edit}      & 1.2  & 1   & 0.3   &                             &           \\
    \addlinespace
    \midrule
    \addlinespace
    Text-to-Video Generation               & VideoCrafter2~\cite{chen2024videocrafter2}       & 1.5  & 1   & 0.9   & Energy-based               & N/A       \\
    \addlinespace
    \bottomrule
  \end{tabular}
\end{table}

\section{Simple Pytorch Implementation}
\begin{figure}[ht!]
    \centering
    \includegraphics[width=0.98\linewidth]{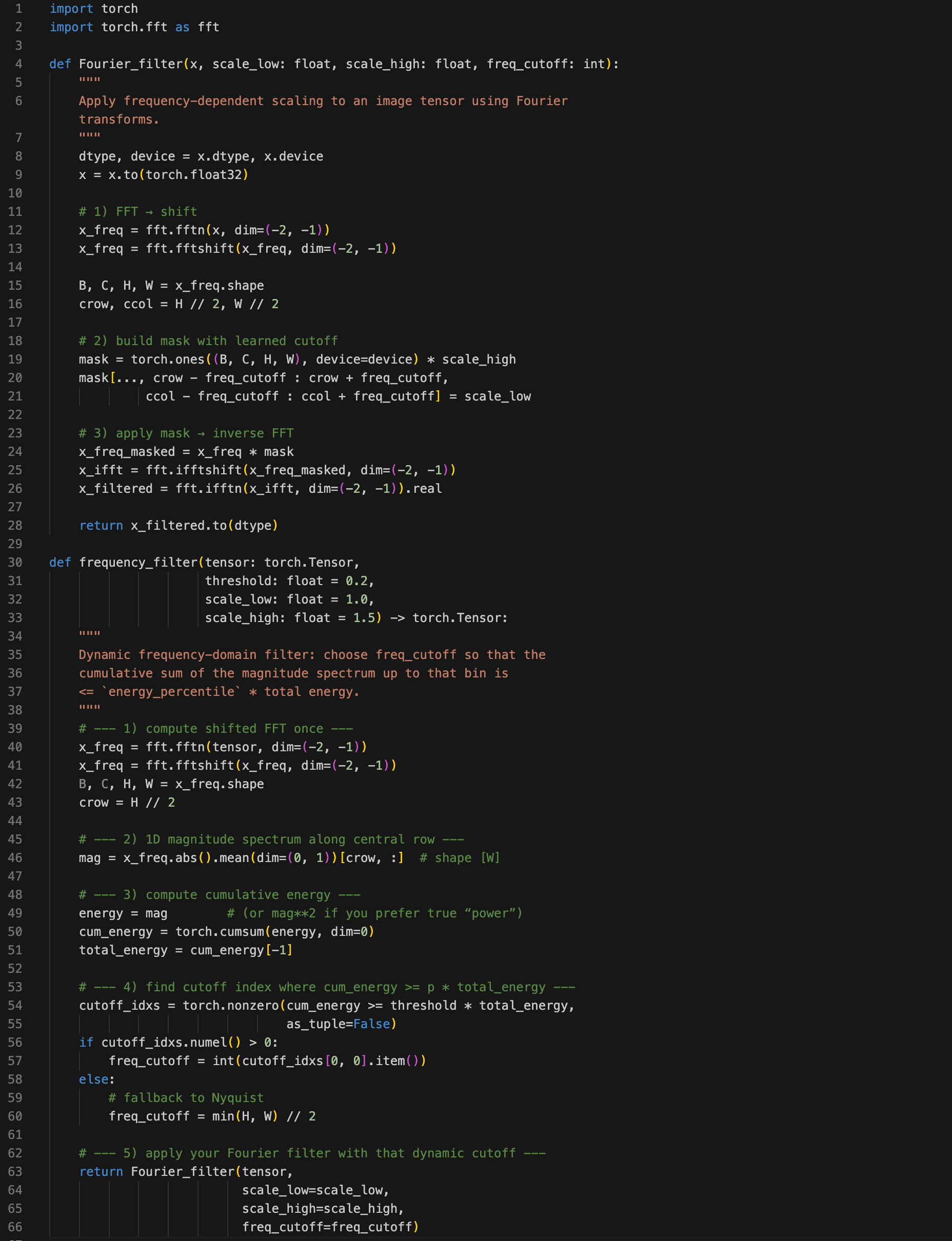}
    \caption{A simple pytorch implementation of our \ourmodel in less than 70 lines of code.}
    \label{fig:pytorch}
\end{figure}

Please refer to \cref{fig:pytorch} for an example implementation of \ourmodel with energy-based cutoff method. 

\end{document}